\documentclass{article}

\PassOptionsToPackage{numbers, compress}{natbib}
\usepackage[final]{neurips_2018}

\usepackage[utf8]{inputenc}
\usepackage[T1]{fontenc}
\usepackage{hyperref}
\usepackage{url}
\usepackage{booktabs}
\usepackage{tikz}

\usepackage{amsmath, amssymb, mathtools}
\usepackage{algorithm}
\usepackage{algpseudocode}
\usepackage{wrapfig}

\usepackage{amsfonts}
\usepackage{nicefrac}
\usepackage{microtype}
\usepackage{pdflscape}
\usepackage{tikz}

\usepackage[toc,page]{appendix}
\usepackage{subfigure}
\usepackage{graphicx}
\usepackage{multirow}
\usepackage{color}
\usepackage{soul}

\allowdisplaybreaks

\newcommand{\minibatch}{\mathcal{M}}

\newcommand{\dkls}[3]{\mathbb{D}_{KL}^{#1}[#2 \, \|\, #3]}

\newcommand\cut[1]{}

\newcommand{\squishlist}{
   \begin{list}{$\bullet$}
    { \setlength{\itemsep}{0pt}      \setlength{\parsep}{3pt}
      \setlength{\topsep}{3pt}       \setlength{\partopsep}{0pt}
      \setlength{\leftmargin}{1.5em} \setlength{\labelwidth}{1em}
      \setlength{\labelsep}{0.5em} } }

\newcommand{\squishlisttwo}{
   \begin{list}{$\bullet$}
    { \setlength{\itemsep}{0pt}    \setlength{\parsep}{0pt}
      \setlength{\topsep}{0pt}     \setlength{\partopsep}{0pt}
      \setlength{\leftmargin}{2em} \setlength{\labelwidth}{1.5em}
      \setlength{\labelsep}{0.5em} } }

\newcommand{\squishend}{
    \end{list}  }

{}
{}
{}
{}
{}
{}

\newcommand{\real}{\mbox{$\mathbb{R}$}}

\newcommand{\sqr}[1]{\left[#1\right]}

\newcommand{\myexpect}{\mathbb{E}}

\newcommand{\gauss}{\mbox{${\cal N}$}}

\newcommand{\myvec}[1]{\mbox{$\mathbf{#1}$}}
\newcommand{\myvecsym}[1]{\mbox{$\boldsymbol{#1}$}}

\newcommand{\vone}{\mbox{$\myvecsym{1}$}}

\newcommand{\vepsilon}{\mbox{$\myvecsym{\epsilon}$}}

\newcommand{\vmu}{\mbox{$\myvecsym{\mu}$}}

\newcommand{\vLambda}{\mbox{$\myvecsym{\Lambda}$}}

\newcommand{\vtheta}{\mbox{$\myvecsym{\theta}$}}

\newcommand{\vSigma}{\mbox{$\myvecsym{\Sigma}$}}

\newcommand{\vd}{\mbox{$\myvec{d}$}}

\newcommand{\vg}{\mbox{$\myvec{g}$}}

\newcommand{\vu}{\mbox{$\myvec{u}$}}
\newcommand{\vv}{\mbox{$\myvec{v}$}}

\newcommand{\vx}{\mbox{$\myvec{x}$}}

\newcommand{\vy}{\mbox{$\myvec{y}$}}

\newcommand{\vA}{\mbox{$\myvec{A}$}}
\newcommand{\vB}{\mbox{$\myvec{B}$}}
\newcommand{\vC}{\mbox{$\myvec{C}$}}
\newcommand{\vD}{\mbox{$\myvec{D}$}}

\newcommand{\vG}{\mbox{$\myvec{G}$}}
\newcommand{\vH}{\mbox{$\myvec{H}$}}
\newcommand{\vI}{\mbox{$\myvec{I}$}}

\newcommand{\vK}{\mbox{$\myvec{K}$}}

\newcommand{\vQ}{\mbox{$\myvec{Q}$}}

\newcommand{\vU}{\mbox{$\myvec{U}$}}
\newcommand{\vV}{\mbox{$\myvec{V}$}}
\newcommand{\vW}{\mbox{$\myvec{W}$}}

\newcommand{\diag}{\mbox{$\mbox{diag}$}}

\newcommand{\calD}{\mbox{${\cal D}$}}

\newcommand{\data}{\calD}

\newcommand{\be}{\begin{equation}}
\newcommand{\ee}{\end{equation}}
\newcommand{\bea}{\begin{eqnarray}}
\newcommand{\eea}{\end{eqnarray}}
\newcommand{\beaa}{\begin{eqnarray*}}
\newcommand{\eeaa}{\end{eqnarray*}}

\title{SLANG: Fast Structured Covariance Approximations for Bayesian Deep Learning with Natural Gradient}

\author{
	Aaron Mishkin\thanks{Equal contributions. This work was conducted during an internship at the RIKEN Center for AI project.}\\
	University of British Columbia\\
	Vancouver, Canada\\
	\texttt{amishkin@cs.ubc.ca}
	\And
	Frederik Kunstner\footnotemark[1] \\
	Ecole Polytechnique Fédérale de Lausanne\\
	Lausanne, Switzerland\\
	\texttt{frederik.kunstner@epfl.ch}
	\And
	Didrik Nielsen\\
	RIKEN Center for AI Project\\
	Tokyo, Japan\\
	\texttt{didrik.nielsen@riken.jp}
	\And
	Mark Schmidt\\
	University of British Columbia\\
	Vancouver, Canada\\
	\texttt{schmidtm@cs.ubc.ca}
	\And
	Mohammad Emtiyaz Khan\\
	RIKEN Center for AI Project\\
	Tokyo, Japan\\
	\texttt{emtiyaz.khan@riken.jp}
}

\begin{document}

\maketitle

\begin{abstract}
   Uncertainty estimation in large deep-learning models is a computationally challenging task, where it is difficult to form even a Gaussian approximation to the posterior distribution.
   In such situations, existing methods usually resort to a diagonal approximation of the covariance matrix despite the fact that these matrices are known to result in poor uncertainty estimates.
   To address this issue, we propose a new stochastic, low-rank, approximate natural-gradient (SLANG) method for variational inference in large, deep models.
   Our method estimates a ``diagonal plus low-rank'' structure based solely on back-propagated gradients of the network log-likelihood.
   This requires strictly less gradient computations than methods that compute the gradient of the whole variational objective.
   Empirical evaluations on standard benchmarks confirm that SLANG enables faster and more accurate estimation of uncertainty than mean-field methods, and performs comparably to state-of-the-art methods.
\end{abstract}


\section{Introduction}

Deep learning has had enormous recent success in fields such as speech recognition and computer vision.
In these problems, our goal is to predict well and we are typically less interested in the uncertainty behind the predictions.
However, deep learning is now becoming increasingly popular in applications such as robotics and medical diagnostics, where accurate measures of uncertainty are crucial for reliable decisions.
For example, uncertainty estimates are important for physicians who use automated diagnosis systems to choose effective and safe treatment options.
Lack of such estimates may lead to decisions that have disastrous consequences.

The goal of Bayesian deep learning is to provide uncertainty estimates by integrating over the posterior distribution of the parameters.
Unfortunately, the complexity of deep learning models makes it infeasible to perform the integration exactly.
Sampling methods such as stochastic-gradient Markov chain Monte Carlo \cite{chen2014stochastic} have been applied to deep models, but they usually converge slowly.
They also require a large memory to store the samples and often need large preconditioners to mix well ~\cite{ahn2012bayesian, balan2015bayesian, simsekli2016stochastic}.
In contrast, variational inference (VI) methods require much less memory and can scale to large problems by exploiting stochastic gradient methods \citep{blundell2015weight, graves2011practical, ranganath2013black}.
However, they often make crude simplifications, like the mean-field approximation, to reduce the memory and computation cost.
This can result in poor uncertainty estimates \cite{turner2011two}.
Fast and accurate estimation of uncertainty for large models remains a challenging problem in Bayesian deep learning.

\begin{figure}[!t]
	\center
	\includegraphics[width=5in]{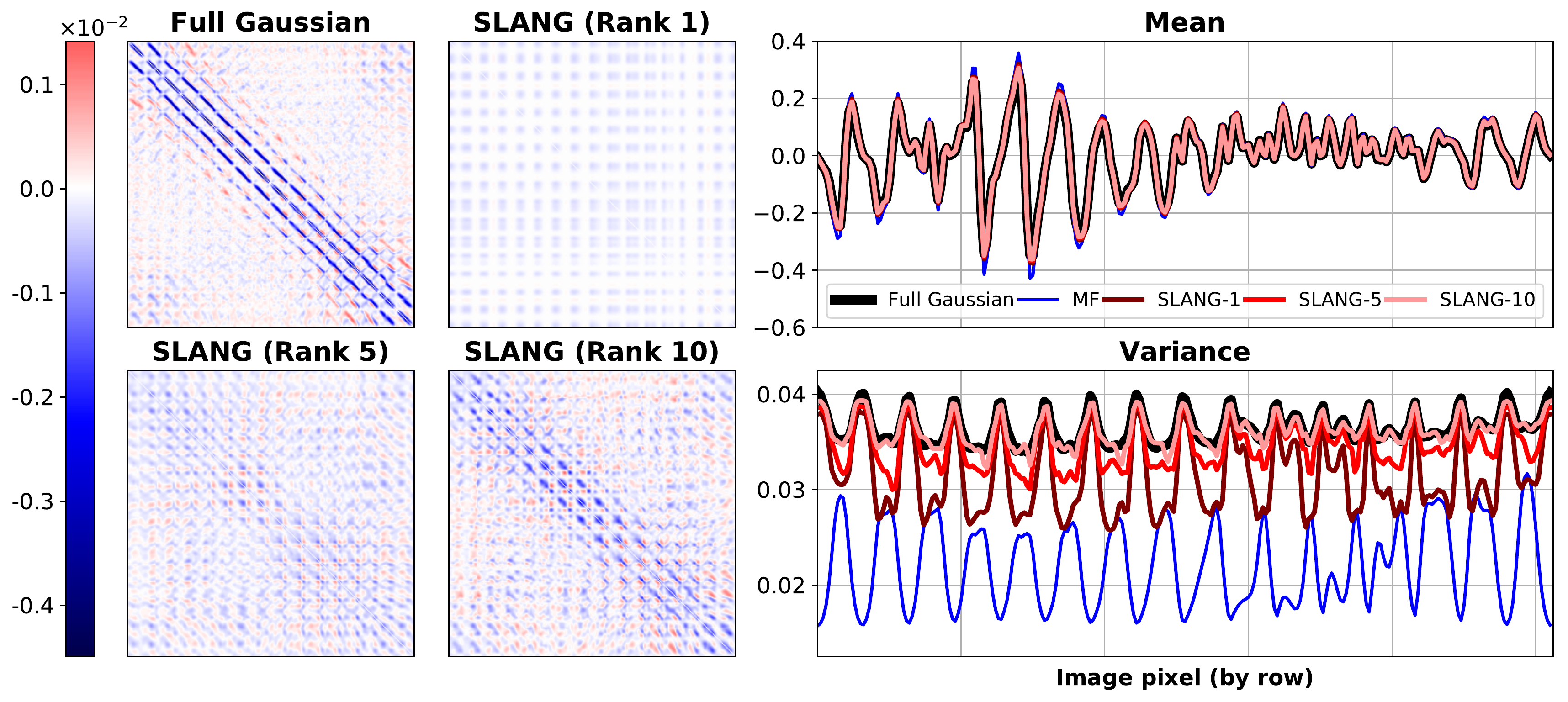}
	\label{fig:1}
	\caption{
This figure illustrates the advantages of SLANG method over mean-field approaches on the USPS dataset (see Section \ref{sec:log_reg} for experimental details).
The figure on the left compares our structured covariance approximation with the one obtained by a full Gaussian approximation. For clarity, only off-diagonal entries are shown. We clearly see that our approximation becomes more accurate as the rank is increased.
The figures on the right compare the means and variances (the diagonal of the covariance).
The means match closely for all methods, but the variance is heavily underestimated by the mean-field method.
SLANG's covariance approximations do not suffer form this problem, which is likely due to the off-diagonal structure it learns.
	}
\end{figure}

In this paper, we propose a new variational inference method to estimate Gaussian approximations with a diagonal plus low-rank covariance structure. This gives more accurate and flexible approximations than the mean-field approach. Our method also enables fast estimation by using an approximate natural-gradient algorithm that builds the covariance estimate solely based on the back-propagated gradients of the network log-likelihood. We call our method \emph{stochastic low-rank approximate
natural-gradient} (SLANG).
SLANG requires strictly less gradient computations than methods that require gradients of the variational objective obtained using the reparameterization trick \cite{miller2017variational, ong2017gaussian, titsias2014doubly}.
Our empirical comparisons demonstrate the improvements obtained over mean-field methods (see Figure \ref{fig:1} for an example) and show that SLANG gives comparable results to the state-of-the-art on standard benchmarks.

The code to reproduce the experimental results in this paper is available at \url{https://github.com/aaronpmishkin/SLANG}.

\subsection{Related Work}
Gaussian variational distributions with full covariance matrices have been used extensively for shallow models
\cite{barber1998ensemble, Challis:11, Jaakkola96b, marlin2011piecewise, miller2017variational, seeger2000bayesian, tan2018gaussian, titsias2014doubly}.
Several efficient ways of computing the full covariance matrix are discussed by Seeger~\cite{seeger2010gaussian}.
Other works have considered various structured covariance approximations, based on the Cholesky decomposition \cite{Challis:11, titsias2014doubly}, sparse covariance matrices \cite{tan2018gaussian} and low-rank plus diagonal structure \cite{barber1998ensemble, ong2017gaussian, seeger2000bayesian}.
Recently, several works \cite{miller2017variational, ong2017gaussian} have applied stochastic gradient descent on the variational objective to estimate such a structure.
These methods often employ an adaptive learning rate method, such as Adam or RMSprop, which increases the memory cost.
All of these methods have only been applied to shallow models, and it remains unclear how they will perform (and whether they can be adapted) for deep models.
Moreover, a natural-gradient method is preferable to gradient-based methods when optimizing the parameters of a distribution ~\cite{amari1998natural, hoffman2013stochastic, khan2018fast}.
Our work shows that a natural-gradient method not only has better convergence properties, but also has lower computation and memory cost than gradient-based methods.

For deep models, a variety of methods have been proposed based on mean-field approximations.
These methods optimize the variational objective using stochastic-gradient methods and differ from each other in the way they compute those gradients~\citep{blundell2015weight, graves2011practical, hernandez15pbp, khan2017vprop, ranganath2013black, zhang2017noisy}.
They all give poor uncertainty estimates in the presence of strong posterior correlations and also shrink variances \cite{turner2011two}.
SLANG is designed to add extra covariance structure and ensure better performance than mean-field approaches.

A few recent works have explored structured covariance approximations for deep models.
In \cite{zhang2017noisy}, the Kronecker-factored approximate curvature (K-FAC) method is applied to perform approximate natural-gradient VI.
Another recent work has applied K-FAC to find a Laplace approximation~\cite{ritter2018scalable}.
However, the Laplace approximation can perform worse than variational inference in many scenarios, e.g., when the posterior distribution is not symmetric \cite{nickisch2008approximations}.
Other types of approximation methods include Bayesian dropout \cite{yarin16dropout} and methods that use matrix-variate Gaussians \cite{louizos2016structured, sun2017learning}.
All of these approaches make structural assumptions that are different from our low-rank plus diagonal structure.
However, similarly to our work, they provide new ways to improve the speed and accuracy of uncertainty estimation in deep learning.


\section{Gaussian Approximation with Natural-Gradient Variational Inference}
Our goal is to estimate the uncertainty in deep models using Bayesian inference.
Given $N$ data examples $\data = \{ \data_i \}_{i=1}^N$, a Bayesian version of a deep model can be specified by using a likelihood $p(\data_i|\vtheta)$ parametrized by a deep network with parameters $\vtheta\in\real^D$ and a prior distribution $p(\vtheta)$.
For simplicity, we assume that the prior is a Gaussian distribution, such as an isotropic Gaussian $p(\vtheta) \sim \gauss(0, (1/\lambda)\vI)$ with the scalar precision parameter $\lambda>0$.
However, the methods presented in this paper can easily be modified to handle many other types of prior distributions.
Given such a model, Bayesian approaches compute an estimate of uncertainty by using the posterior distribution: $p(\vtheta | \data) = p(\data|\vtheta) p(\vtheta)/p(\data)$.
This requires computation of the \emph{marginal likelihood} $p(\data) = \int p(\data|\vtheta) p(\vtheta) d\vtheta$, which is a high-dimensional integral and difficult to compute.

Variational inference (VI) simplifies the problem by approximating $p(\vtheta | \data)$ with a distribution $q(\vtheta)$.
In this paper, our focus is on obtaining approximations that have a Gaussian form, i.e., $q(\vtheta) = \gauss(\vtheta|\vmu,\vSigma)$ with mean $\vmu$ and covariance $\vSigma$.
The parameters $\vmu$ and $\vSigma$ are referred to as the \emph{variational parameters} and can be obtained by maximizing a lower bound on $p(\data)$ called the evidence lower bound (ELBO),
   \begin{align}
	\textrm{ELBO: } \quad  \mathcal{L}(\vmu,\vSigma) := \myexpect_q \sqr{\log p(\data|\vtheta)} - \dkls{}{q(\vtheta)}{p(\vtheta)}. \label{eq:elbo}
	\end{align}
where $\mathbb{D}_{KL}[\cdot]$ denotes the Kullback-Leibler divergence.

A straightforward and popular approach to optimize $\mathcal{L}$ is to use stochastic gradient methods \cite{miller2017variational, ong2017gaussian, ranganath2013black, titsias2014doubly}.
However, natural-gradients are preferable when optimizing the parameters of a distribution \cite{amari1998natural, hoffman2013stochastic, khan2018fast}.
This is because natural-gradient methods perform optimization on the Riemannian manifold of the parameters, which can lead to a faster convergence when compared to gradient-based methods.
Typically, natural-gradient methods are difficult to implement, but many easy-to-implement updates have been derived in recent works \cite{hoffman2013stochastic, khan2017conjugate, khan2017vprop, zhang2017noisy}.
We build upon the approximate natural-gradient method proposed in \cite{khan2017vprop} and modify it to estimate structured covariance-approximations.

Specifically, we extend the Variational Online Gauss-Newton (VOGN) method \cite{khan2017vprop}.
This method uses the following update for $\vmu$ and $\vSigma$ (a derivation is in Appendix \ref{app:vogn_deriv}),
\begin{align}
   \vmu_{t+1} &= \vmu_t - \alpha_t \vSigma_{t+1} \sqr{\hat{\vg}(\vtheta_t) + \lambda \vmu_t}, \quad  \vSigma^{-1}_{t+1} = (1- \beta_t) \vSigma^{-1}_{t}+ \beta_t \sqr{\hat{\vG}(\vtheta_t) + \lambda \vI},  \label{eq:von}\\
              &\quad\quad\textrm{ with } \hat{\vg}(\vtheta_t) := - \frac{N}{M} \sum_{i\in\minibatch} \vg_i(\vtheta_t), \,\, \textrm{ and } \hat{\vG}(\vtheta_t) := - \frac{N}{M}\sum_{i\in\minibatch} \vg_i(\vtheta_t) \vg_i(\vtheta_t)^\top,
\nonumber
\end{align}
where $t$ is the iteration number, $\alpha_t, \beta_t>0$ are learning rates, $\vtheta_t\sim \gauss(\vtheta|\vmu_t, \vSigma_t)$, $\vg_i(\vtheta_t) := \nabla_\theta \log p(\data_i|\vtheta_t)$ is the back-propagated gradient obtained on the $i$'th data example, $\hat{\vG}(\vtheta_t)$ is an Empirical Fisher (EF) matrix, and $\minibatch$ is a minibatch of $M$ data examples.
This update is an approximate natural-gradient update obtained by using the EF matrix as an approximation of the Hessian \cite{martens2014new} in a method called Variational Online Newton (VON) \cite{khan2017vprop}. This is explained in more detail in Appendix \ref{app:vogn_deriv}.
As discussed in \cite{khan2017vprop}, the VOGN method is an
approximate Natural-gradient update which may not have the same properties as the exact natural-gradient update.
However, an advantage of the update \eqref{eq:von} is that it only requires back-propagated gradients, which is a desirable feature when working with deep models.

The update \eqref{eq:von} is computationally infeasible for large deep models because it requires the storage and inversion of the $D\times D$ covariance matrix.
Storage takes $O(D^2)$ memory space and inversion requires $O(D^3)$ computations, which makes the update very costly to perform for large models.
We cannot form $\vSigma$ or invert it when $D$ is in millions.
Mean-field approximations avoid this issue by restricting $\vSigma$ to be a diagonal matrix, but they often give poor Gaussian approximations.
Our idea is to estimate a low-rank plus diagonal approximation of $\vSigma$ that reduces the computational cost while preserving some off-diagonal covariance structure.
In the next section, we propose modifications to the update \eqref{eq:von} to obtain a method whose time and space complexity are both linear in $D$.


\section{Stochastic, Low-rank, Approximate Natural-Gradient (SLANG) Method}
\label{sec:SLANG}

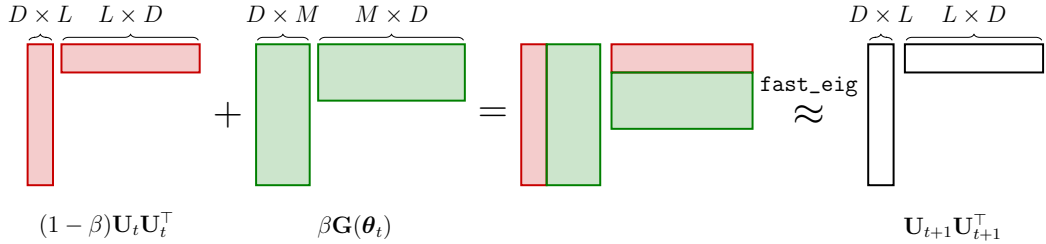
\begin{figure}[!t]
	\centering
	\hspace{-4.6in}
	\makebox[\textwidth][c]{\scalebox{.75}{\usetikzlibrary{decorations.pathreplacing}

\newif\ifmodeone
\modeonetrue
\begin{tikzpicture}

\newcommand{\fontscaling}{\huge}
\newcommand{\fontscalingOne}{\large}
\definecolor{darkred}{RGB}{200,0,0}
\definecolor{darkgreen}{RGB}{0,128,0}
\newcommand{\colorA}{\color{darkred}}
\newcommand{\colorB}{\color{darkgreen}}

\tikzstyle{rectanglestyle} = [line width=1]
\tikzstyle{redrect} = [draw opacity=1, fill opacity=0.2, color=darkred, fill]
\tikzstyle{greenrect} = [draw opacity=1, fill opacity=0.2, color=darkgreen, fill]
\tikzstyle{braces} = [decorate,decoration={brace, amplitude=3pt, raise=2pt},yshift=0pt]
\tikzstyle{bracestwo} = [decorate,decoration={brace, amplitude=3pt, raise=2pt},yshift=0pt]
\tikzstyle{bracestext} = [midway, yshift=15pt,rotate=0]
\tikzstyle{leftbraces} = [decorate,decoration={brace, amplitude=5pt, raise=2pt},yshift=0pt]
\tikzstyle{leftbracestext} = [midway, xshift=-15pt,rotate=90]
\newcommand{\rectanglestyle}{line width=5}

\draw [rectanglestyle, redrect] (-3.25,2) rectangle (-2.8,-0.5);
\draw [rectanglestyle, redrect] (-2.65,2) rectangle (-0.2,1.5) ;

\node at (0.3,0.75) {\fontscaling$+$};

\draw [rectanglestyle, greenrect] (0.8,2) rectangle (1.75,-0.5);
\draw [rectanglestyle, greenrect] (1.9,2) rectangle (4.5,1) ;

\node at (5,0.75) {\fontscaling$=$};

\draw [rectanglestyle, redrect] (5.95,-0.5) rectangle (5.5,2);
\draw [rectanglestyle, redrect] (7.1,2) rectangle (9.6,1.5) ;
\draw [rectanglestyle, greenrect] (5.95,2) rectangle (6.9,-0.5);
\draw [rectanglestyle, greenrect] (7.1,1.5) rectangle (9.6,0.5) ;

\node at (10.6,0.75) {\fontscaling$\approx$};

\draw [rectanglestyle] (11.65,2) rectangle (12.1,-0.5);
\draw [rectanglestyle] (12.3,2) rectangle (14.75,1.5) ;

\node at (-1.85,-1.2) {\fontscalingOne $(1-\beta)\mathbf{U}_t\mathbf{U}_t^\top$};
\node at (2.55,-1.25) {\fontscalingOne $\beta\mathbf{G}(\vtheta_t)$};
\node at (10.6,1.25) {\fontscalingOne \texttt{fast\_eig}};
\node at (13.15,-1.25) {\fontscalingOne $\mathbf{U}_{t+1}\mathbf{U}_{t+1}^\top$};
{
\node (v1) at (-2.75,2) {};
\node (v2) at (-0.1,2) {};
\node (v1_1) at (1.8,2) {};
\node (v2_1) at (4.6,2) {};
\node (v1_2) at (7.1,2) {};
\node (v2_2) at (9.6,2) {};
\node (v2_3) at (14.85,2) {};
\node (v1_3) at (12.2,2) {};

\draw [braces] (v1) -- (v2) node [bracestext] {\fontscalingOne $L \times D$};
\draw [braces] (v1_1) -- (v2_1) node [bracestext] {\fontscalingOne $M \times D$};
\draw [braces] (v1_3) -- (v2_3) node [bracestext] {\fontscalingOne $L \times D$};
}
{
\node (v1) at (-3.35,2) {};
\node (v2) at (-2.7,2) {};
\node (v1_1) at (0.7,2) {};
\node (v2_1) at (1.85,2) {};
\node (v1_2) at (5.4,2) {};
\node (v2_2) at (7,2) {};
\node (v2_3) at (12.2,2) {};
\node (v1_3) at (11.55,2) {};

\draw [braces] (v1) -- (v2) node [bracestext] {\fontscalingOne $D \times L$};
\draw [braces] (v1_1) -- (v2_1) node [bracestext] {\fontscalingOne $D \times M$};
\draw [braces] (v1_3) -- (v2_3) node [bracestext] {\fontscalingOne $D \times L$};
}

\clip (-4.0, 3.0) rectangle (15.0, -2.0);

%
%
%
%

\end{tikzpicture}}}
	\caption{This figure illustrates Equations \eqref{eq:deriv_update_2} and \eqref{eq:low_rank_update} which are used to derive SLANG.}
	\label{fig:visualization}
\end{figure}

Our goal is to modify the update \eqref{eq:von} to obtain a method whose time and space complexity is linear in $D$.
We propose to approximate the inverse of $\vSigma_t$ by a ``low-rank plus diagonal'' matrix:
\begin{align}
	\label{eq:structure}
	\vSigma^{-1}_t \approx \hat{\vSigma}{}_t^{-1} := \vU_t\vU_t^\top + \vD_t,
\end{align}
where $\vU_t$ is a $D\times L$ matrix with $L \ll D$ and $\vD_t$ is a $D\times D$ diagonal matrix.
The cost of storing and inverting this matrix is linear in $D$ and reasonable when $L$ is small.
We now derive an update for $\vU_t$ and $\vD_t$
such that the resulting $\hat{\vSigma}{}_{t+1}^{-1}$ closely approximates the update shown in \eqref{eq:von}.
We start by writing an approximation to the update of $\vSigma_{t+1}^{-1}$ where we replace covariance matrices by their structured approximations:
\begin{align}
   \hat{\vSigma}{}^{-1}_{t+1} := \vU_{t+1}\vU_{t+1}^\top + \vD_{t+1} \approx (1- \beta_t) \hat{\vSigma}{}_t^{-1} + \beta_t \sqr{\hat{\vG}(\vtheta_t) + \lambda \vI}
   \label{eq:approx_update_cov}
\end{align}
This update cannot be performed exactly without potentially increasing the rank of the low-rank component $\vU_{t+1}$, since the structured components on the right hand side are of rank at most $L + M$, where $M$ is the size of the minibatch. This is shown in \eqref{eq:deriv_update_1} below where we have rearranged the left hand side of \eqref{eq:approx_update_cov} as the sum of a structured component and a diagonal component.
To obtain a rank $L$ approximation to the left hand side of \eqref{eq:deriv_update_1}, we propose to approximate the structured component by an eigenvalue decomposition as shown in \eqref{eq:deriv_update_2} below,
\begin{align}
	\label{eq:deriv_update_1}
	(1-\beta_t) \hat{\vSigma}{}_t^{-1}
	+ \beta_t \sqr{\hat{\vG}(\vtheta_t) + \lambda \vI}
	&=
	\underbrace{ (1-\beta_t) \vU_t \vU_t^\top + \beta_t \hat{\vG}(\vtheta_t) }_{\textrm{Rank at most $L+M$}}
	&&+ \underbrace{ (1-\beta_t)\vD_t + \beta_t \lambda \vI }_{\textrm{Diagonal component}},
	\\
	\label{eq:deriv_update_2}
	&\approx
	\quad \quad \ \underbrace{\vQ_{1:L} \vLambda_{1:L} \vQ_{1:L}^\top }_{\textrm{Rank $L$ approximation}}
	&&+
	\underbrace{ (1-\beta_t)\vD_t + \beta_t \lambda \vI }_{\textrm{Diagonal component}},
\end{align}
where $\vQ_{1:L}$ is a $D\times L$ matrix containing the first $L$ leading eigenvectors of $(1-\beta_t) \vU_t \vU_t^\top + \beta_t \hat{\vG}(\vtheta_t)$ and $\vLambda_{1:L}$ is an $L\times L$ diagonal matrix containing
the corresponding eigenvalues. Figure \ref{fig:visualization} visualizes the update from \eqref{eq:deriv_update_1} to \eqref{eq:deriv_update_2}.

The low-rank component $\vU_{t+1}$ can now be updated to mirror the low-rank component of \eqref{eq:deriv_update_2},
\begin{align}
\vU_{t+1} = \vQ_{1:L}\vLambda_{1:L}^{1/2},  \label{eq:low_rank_update},
\end{align}
and the diagonal $\vD_{t+1}$ can be updated to match the diagonal of the left and right sides of \eqref{eq:approx_update_cov}, i.e.,
\begin{align}
   \diag\sqr{\vU_{t+1}\vU_{t+1}^\top + \vD_{t+1} } = \diag\sqr{(1-\beta_t) \vU_t \vU_t^\top + \beta_t \hat{\vG}(\vtheta_t)
   +  (1-\beta_t)\vD_t + \beta_t \lambda \vI },
\end{align}
This gives us the following update for $\vD_{t+1}$ using a \emph{diagonal correction} $\Delta_t$,
\begin{align}
	\label{eq:slang_diagonal_update}
	\vD_{t+1} &= (1-\beta) \vD_t + \beta_t \lambda \vI + \Delta_t,\\
	\label{eq:slang_diagonal_difference}
	\Delta_t &= \text{diag} \left[(1-\beta)\vU_{t}\vU_{t}^\top + \beta_t \hat{\vG}(\vtheta_t) - \vU_{t+1} \vU_{t+1}^\top \right].
\end{align}
This step is cheap since computing the diagonal of the EF matrix is linear in $D$.

The new covariance approximation can now be used to update $\vmu_{t+1}$ according to \eqref{eq:von} as shown below:
\begin{align} \label{eq:slang_final_update}
   \textrm{SLANG: } \quad \vmu_{t+1} &= \vmu_t - \alpha_t \sqr{ \vU_{t+1}\vU_{t+1}^\top + \vD_{t+1} }^{-1} \sqr{\hat{\vg}(\vtheta_t) + \lambda \vmu_t},
\end{align}
The above update uses a stochastic, low-rank covariance estimate to approximate natural-gradient updates, which is why we use the name SLANG.

When $L = D$, $\vU_{t+1} \vU_{t+1}^\top$ is full rank and SLANG is identical to the approximate natural-gradient update \eqref{eq:von}.
When $L < D$, SLANG produces matrices $\hat{\vSigma}{}_t^{-1}$ with diagonals matching \eqref{eq:von} at every iteration.
The diagonal correction ensures that no diagonal information is lost during the low-rank approximation of the covariance.
A formal statement and proof is given in Appendix~\ref{:appendix_diagonal_correction}.

We also tried an alternative method where $\vU_{t+1}$ is learned using an exponential moving-average of the eigendecompositions of $\hat{\vG}(\vtheta)$.
This previous iteration of SLANG is discussed in Appendix~\ref{appendix:old_slang}, where we show that it gives worse results than the SLANG update.

Next, we give implementation details of SLANG.


\subsection{Details of the SLANG Implementation}
\label{sec:alg_details}
The pseudo-code for SLANG is shown in Algorithm \ref{alg:slang} in Figure \ref{fig:slang}.

At every iteration, we generate a sample $\vtheta_t \sim \gauss(\vtheta|\vmu_t, \vU_t\vU_t^\top + \vD_t)$.
This is implemented in line 4 of Algorithm \ref{alg:slang} using the subroutine {\tt fast\_sample}.
Pseudocode for this subroutine is given in Algorithm \ref{alg:fast-sample}.
This function uses the Woodbury identity and to compute the square-root matrix $\vA_t = \big(\vU_t\vU_t^\top + \vD_t\big)^{-1/2}$ ~\cite{ambikasaran2014fast}.
The sample is then computed as $\vtheta_t = \vmu_t + \vA_t \vepsilon$, where $\vepsilon \sim \mathcal{N}(\mathbf{0}, \vI)$.
The function {\tt fast\_sample} requires computations in $O(DL^2 + DLS)$ to generate $S$ samples, which is linear in $D$.
More details are given in Appendix \ref{apx:fast-sample}.

Given a sample, we need to compute and store all the individual stochastic gradients $\vg_i(\vtheta_t)$ for all examples $i$ in a minibatch $\minibatch$. The standard back-propagation implementation does not allow this.
We instead use a version of the backpropagation algorithm outlined in a note by Goodfellow \cite{goodfellow2015efficient}, which enables efficient computation of the gradients $\hat{\vg}_i(\vtheta_t)$.
This is shown in line 6 of Algorithm \ref{alg:slang}, where a subroutine \texttt{backprop\_goodfellow} is used (see details of this subroutine in Appendix \ref{apx:backprop-goodfellow}).

In line 7, we compute the eigenvalue decomposition of $ (1-\beta_t) \vU_t \vU_t + \beta_t \hat{\vG}(\vtheta_t)$ by using the \texttt{fast\_eig} subroutine.
The subroutine \texttt{fast\_eig} implements a randomized eigenvalue decomposition method discussed in \cite{halko2011finding}.
It computes the top-$L$ eigendecomposition of a low-rank matrix in $O(DLMS + DL^2)$.
More details on the subroutine are given in Appendix \ref{apx:eigendecomposition}.
The matrix $\vU_{t+1}$ and $\vD_{t+1}$  are updated using the eigenvalue decomposition in lines 8, 9 and 10.

In lines 11 and 12, we compute the update vector $[\vU_{t+1}\vU_{t+1}^\top + \vD_{t+1} ]^{-1} \sqr{\hat{\vg}(\vtheta_t) + \lambda \vmu_t}$, which requires solving a linear system.
We use the subroutine \texttt{fast\_inverse} shown in Algorithm \ref{alg:fast-inverse}.
This subroutine uses the Woodbury identity to efficiently compute the inverse with a cost $O(DL^2)$.
More details are given in Appendix \ref{apx:fast-inverse}.
Finally, in line 13, we update $\vmu_{t+1}$.

The overall computational complexity of SLANG is $O(DL^2 + DLMS)$ and its memory cost is $O(DL + DMS)$.
Both are linear in $D$ and $M$.
The cost is quadratic in $L$, but since $L \ll D$ (e.g., 5 or 10), this only adds a small multiplicative constant in the runtime.
SLANG reduces the cost of the update \eqref{eq:von} significantly while preserving some posterior correlations.

\begin{figure*}[ttt!]
	\begin{minipage}[t]{3.2in}
		\begin{algorithm}[H]
			\caption{\texttt{SLANG}\protect\phantom{\texttt{fast\_sample}$(\vmu, \vU, \vd)$}}
			\begin{algorithmic}[1]
				\Require Data $\data$, hyperparameters $M, L, \lambda, \alpha, \beta$
				\State Initialize $\vmu, \vU ,\vd$
				\State $\delta \gets (1- \beta)$
				\While{not converged}
					\State $\vtheta \gets \texttt{fast\_sample}(\vmu,\vU,\vd)$
					\State $\minibatch \gets$ sample a minibatch
					\State $[\vg_1,..,\vg_M] \gets \texttt{backprop\_goodfellow}(\data_\minibatch, \vtheta)$
					\State $\vV \gets \texttt{fast\_eig}(\delta\vu_1,..,\delta\vu_L,\beta\vg_1,..,\beta\vg_M, L)$
					\State $\Delta_d \gets \sum_{i=1}^L \delta\vu_i^2 + \sum_{i=1}^M \beta\vg_i^2 - \sum_{i=1}^L \vv_i^2$
					\State $\vU \gets \vV$
					\State $\vd \gets \delta\vd + \Delta_d + \lambda\vone$
					\State $\hat{\vg} \gets \sum_i \vg_i + \lambda\vmu$
					\State $\Delta_\mu \gets \texttt{fast\_inverse}(\hat{\vg}, \vU,\vd)$
					\State $\vmu \gets \vmu - \alpha \Delta_\mu$
				\EndWhile
			\State \Return \vmu, \vU, \vd
			\end{algorithmic}
			\label{alg:slang}
		\end{algorithm}
	\end{minipage}
	\hfill
	\begin{minipage}[t]{2.3in}
		\begin{algorithm}[H]
			\begin{algorithmic}[1]
				\State $\vA \gets (\vI_L + \vU^\top \vd^{-1} \vU)^{-1}$
				\State $\vy \gets \vd^{-1} \vg - \vd^{-1} \vU \vA \vU^\top \vd^{-1} \vg$
				\State \Return \vy
			\end{algorithmic}
			\caption{\texttt{fast\_inverse}$(\vg, \vU, \vd)$}
			\label{alg:fast-inverse}
		\end{algorithm}
		\vspace{-.2em}
		\begin{algorithm}[H]
			\begin{algorithmic}[1]
				\State $\vepsilon \sim \mathcal{N}(0, \vI_D)$
				\State $\vV \gets \vd^{-1/2} \odot \vU$
				\State $\vA \gets \text{Cholesky}(\vV^\top \vV)$
				\State $\vB \gets \text{Cholesky}(\vI_L + \vV^\top \vV)$
				\State $\vC \gets \vA^{-\top}(\vB - \vI_L)\vA^{-1}$
				\State $\vK \gets (\vC + \vV^\top \vV)^{-1}$
				\State $\vy \gets \vd^{-1/2}\vepsilon - \vV \vK \vV^\top \vepsilon$
				\State \Return $\vmu + \vy$
			\end{algorithmic}
			\caption{\texttt{fast\_sample}$(\vmu, \vU, \vd)$}
			\label{alg:fast-sample}
		\end{algorithm}
	\end{minipage}
	\caption{
		Algorithm \ref{alg:slang} gives the pseudo-code for SLANG.
		Here, $M$ is the minibatch size, $L$ is the number of low-rank factors, $\lambda$ is the prior precision parameter, and $\alpha, \beta$ are learning rates.
		The diagonal component is denoted with a vector $\vd$ and the columns of the matrix $\vU$ and $\vV$ are denoted by $\vu_j$ and $\vv_j$ respectively.
		The algorithm depends on multiple subroutines, described in more details in Section \ref{sec:alg_details}.
		The overall complexity of the algorithm is $O(DL^2 + DLM)$.
	}
	\label{fig:slang}
\end{figure*}


\section{Experiments}
In this section, our goal is to show experimental results in support of the following claims: (1) SLANG gives reasonable posterior approximations, and (2) SLANG performs well on standard benchmarks for Bayesian neural networks.
We present evaluations on several LIBSVM datasets, the UCI regression benchmark, and MNIST classification.
SLANG beats mean-field methods on almost all tasks considered and performs comparably to state-of-the-art methods. SLANG also converges faster than mean-field methods.

\subsection{Bayesian Logistic Regression}
\label{sec:log_reg}

\begin{table}[]
   \caption{Results on Bayesian logistic regression where we compare SLANG to three full-Gaussian methods and three mean-field methods.
   We measure negative ELBO, test log-loss, and symmetric KL-divergence between each approximation and the Full-Gaussian Exact method (last column).
   Lower values are better.
   SLANG nearly always gives better results than the mean-field methods, and with $L=10$ is comparable to Full-Gaussian methods.
   This shows that our structured covariance
   approximation is reasonably accurate for Bayesian logistic regression.}
\setlength\tabcolsep{3.6pt}
\begin{tabular}{llccccccccccc}
                                                                         &                                                  & \multicolumn{3}{c}{Mean-Field Methods} & \hspace{0.1cm} & \multicolumn{3}{c}{SLANG} & \hspace{0.1cm} & \multicolumn{3}{c}{Full Gaussian} \\ \cmidrule(lr){3-5} \cmidrule(lr){7-9} \cmidrule(lr){11-13}
\textbf{Dataset}                                                         & \textbf{Metrics}                         & \textbf{EF} & \textbf{Hess.} & \textbf{Exact} &  & \textbf{L = 1} & \textbf{L = 5} & \textbf{L = 10}    &  & \textbf{EF} & \textbf{Hess.} & \textbf{Exact} \\ \hline
\multirow{3}{*}{Australian}                                              & ELBO                                             & 0.614      & 0.613        & 0.593      &              & 0.574   & 0.569    & \textbf{0.566}     & & 0.560             & 0.558         & 0.559    \\
                                                                         & NLL                                              & 0.348      & 0.347        & 0.341      &              & 0.342   & 0.339    & \textbf{0.338}     & & 0.340             & 0.339         & 0.338    \\
                                                                         & KL {\small( $\times 10^4$)}                      & 2.240      & 2.030        & 0.195      &              & 0.033   & 0.008    & \textbf{0.002}     & & 0.000             & 0.000         & 0.000    \\ \hline
\multirow{3}{*}{\begin{tabular}[c]{@{}l@{}}Breast\\ Cancer\end{tabular}} & ELBO                                             & 0.122      & 0.121        & 0.121      &              & 0.112   & 0.111    & \textbf{0.111}     & & 0.111             & 0.109         & 0.109    \\
                                                                         & NLL                                              & 0.095      & 0.094        & 0.094      &              & 0.092   & 0.092    & \textbf{0.092}     & & 0.092             & 0.091         & 0.091    \\
                                                                         & KL {\small( $\times 10^0$)}                      & 8.019      & 9.071        & 7.771      &              & 0.911   & 0.842    & \textbf{0.638}     & & 0.637             & 0.002         & 0.000    \\ \hline
\multirow{3}{*}{a1a}                                                     & ELBO                                             & 0.384      & 0.383        & 0.383      &              & 0.377   & 0.374    & \textbf{0.373}     & & 0.369             & 0.368         & 0.368    \\
                                                                         & NLL                                              & 0.339      & 0.339        & 0.339      &              & 0.339   & 0.339    & \textbf{0.339}     & & 0.339             & 0.339         & 0.339    \\
                                                                         & KL {\small($\times 10^2$)}                       & 2.590      & 2.208        & 1.295      &              & 0.305   & 0.173    & \textbf{0.118}     & & 0.014             & 0.000         & 0.000    \\ \hline
\multirow{3}{*}{\begin{tabular}[c]{@{}l@{}}USPS\\ 3vs5\end{tabular}}   & ELBO                                             & 0.268      & 0.268        & 0.267      &              & 0.210   & 0.198    & \textbf{0.193}     & & 0.189             & 0.186         & 0.186    \\
                                                                         & NLL                                              & 0.139      & 0.139        & 0.138      &              & 0.132   & 0.132    & \textbf{0.131}     & & 0.131             & 0.130         & 0.130    \\
                                                                         & KL {\small($\times 10^1$)}                       & 7.684      & 7.188        & 7.083      &              & 1.492   & 0.755    & \textbf{0.448}     & & 0.180             & 0.001         & 0.000
\end{tabular}
\label{tbl:logRegExperiments}
\end{table}

We considered four benchmark datasets for our comparison: USPS 3vs5, Australian, Breast-Cancer, and a1a.
Details of the datasets are in Table \ref{data_stat} in Appendix \ref{app:implement_logreg} along with the implementation details of the methods we compare to.
We use L-BFGS~\cite{wright1999numerical} to compute the optimal full-Gaussian variational approximation that minimizes the ELBO using the method described in Marlin et al.~\cite{marlin2011piecewise}.
We refer to the optimal full-Gaussian variational approximation as the ``Full-Gaussian Exact'' method.
We also compute the optimal mean-field Gaussian approximation and refer to it as ``MF Exact''.

Figure \ref{fig:1} shows a qualitative comparison of the estimated posterior means, variances, and covariances for the USPS-3vs5 dataset ($N=770$, $D=256$).
The figure on the left compares covariance approximations obtained with SLANG to the Full-Gaussian Exact method.
Only off-diagonal entries are shown. We see that the approximation becomes more and more accurate as the rank is increased. The figures on the right compare the means and variances.
The means match closely for all methods, but the variance is heavily underestimated by the MF Exact method;
we see that the variances obtained under the mean-field approximation estimate a high variance where Full-Gaussian Exact has a low variance and vice-versa.
This ``trend-reversal'' is due to the typical shrinking behavior of mean-field methods \cite{turner2011two}.
In contrast, SLANG corrects the trend reversal problem even when $L=1$.
Similar results for other datasets are shown in Figure \ref{fig:plot1-logreg} in Appendix \ref{app:add_logreg}.

The complete results for Bayesian logistic regression are summarized in Table \ref{tbl:logRegExperiments}, where we also compare to four additional methods called ``Full-Gaussian EF'', ``Full-Gaussian Hessian'', ``Mean-Field EF'', and ``Mean-Field Hessian''.
The Full-Gaussian EF method is the natural-gradient update \eqref{eq:von} which uses the EF matrix $\hat{\vG}(\vtheta)$, while the Full-Gaussian Hessian method uses the Hessian instead of the EF matrix (the updates are given in
\eqref{eq:Van_mean_1} and \eqref{eq:Van_var_1} in Appendix \ref{app:vogn_deriv}).
The last two methods are the mean-field versions of the Full-Gaussian EF and Full-Gaussian Hessian methods, respectively.
We compare negative ELBO, test log-loss using cross-entropy, and symmetric KL-divergence between the approximations and the Full-Gaussian Exact method.
We report averages over 20 random 50\%-50\% training-test splits of the dataset.
Variances and results for SLANG with $L=2$ are omitted here due to space constraints, but are reported in Table \ref{tbl:logRegExperiments-baselines} in Appendix \ref{app:add_logreg}.

We find that SLANG with $L=1$ nearly always produces better approximations than the mean-field methods.
As expected, increasing $L$ improves the quality of the variational distribution found by SLANG according to all three metrics.
We also note that Full-Gaussian EF method has similar performance to the Full-Gaussian Hessian method, which indicates that the EF approximation may be acceptable for Bayesian logistic regression.

\begin{figure*}[ttt!]
\centering
\includegraphics[width=\textwidth]{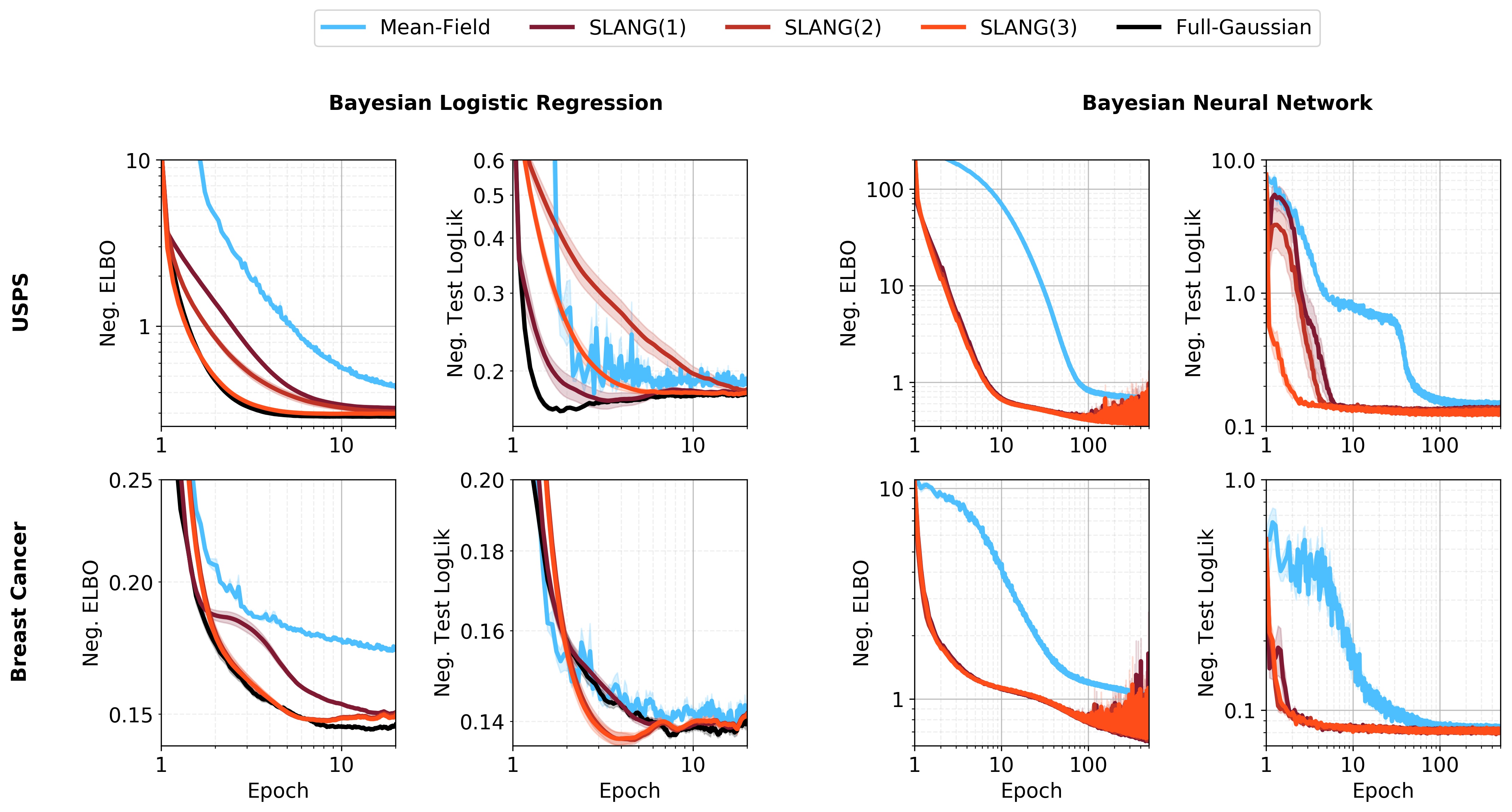}
\caption{
This figure compares the convergence behavior on two datasets: USPS 3vs5 (top) and Breast Cancer (bottom);
and two models: Bayesian logistic regression (left) and Bayesian neural networks (BNN) (right).
The three methods SLANG(1, 2, 3) refer to SLANG with $L=1, 5, 10$ for logistic regression. For BNN, they refer to SLANG with $L=8, 16, 32$.
The mean-field method is a natural-gradient mean-field method for logistic regression (see text) and BBB \cite{blundell2015weight} for BNN.
This comparison clearly shows that SLANG converges faster than the mean-field method, and, for Bayesian logistic regression, matches the convergence of the full-Gaussian method when $L$ is increased.
}
\label{fig:convergence}
\end{figure*}

The left side in Figure \ref{fig:convergence} shows convergence results on the USPS 3vs5 and Breast Cancer datasets.
The three methods SLANG(1, 2, 3) refer to SLANG with $L=1, 5, 10$.
We compare these three SLANG methods to Mean-Field Hessian and Full-Gaussian Hessian.
SLANG converges faster than the mean-field method, and matches the convergence of the full-Gaussian method when $L$ is increased.

\subsection{Bayesian Neural Networks (BNNs)}
\label{sec:uci}

An example for Bayesian Neural Networks on a synthetic regression dataset is given in Appendix \ref{app:toyexample}, where we illustrate the quality of SLANG's posterior covariance.

The right side in Figure \ref{fig:convergence} shows convergence results for the USPS 3vs5 and Breast Cancer datasets.
Here, the three methods SLANG(1, 2, 3) refer to SLANG with $L=8, 16, 32$.
We compare SLANG to a mean-field method called Bayes by Backprop \cite{blundell2015weight}.
Similar to the Bayesian logistic regression experiment, SLANG converges much faster than the mean-field method.
However, the ELBO convergence for SLANG shows that the optimization procedure does not necessarily converge to a local minimum.
This issue does not appear to affect the test log-likelihood.
While it might only be due to stochasticity, it is possible that the problem is exacerbated by the replacement of the Hessian with the EF matrix.
We have not determined the specific cause and it warrants further investigation in future work.

\begin{table}[t]
\setlength{\tabcolsep}{4pt}
\centering
\caption{Comparison on UCI datasets using Bayesian neural networks. We repeat the setup used in Gal and Ghahramani~\cite{yarin16dropout}. SLANG uses $L=1$, and outperforms BBB but gives comparable performance to Dropout.}
\label{table:uci}
\begin{tabular}{l c c c c c c c}
\hline
& \multicolumn{3}{c}{Test RMSE} & \hspace{0.1cm} & \multicolumn{3}{c}{Test log-likelihood} \\
\bf{Dataset} & \bf{BBB} & \bf{Dropout} & \bf{SLANG} & & \bf{BBB} & \bf{Dropout} & \bf{SLANG} \\
\hline
Boston   & 3.43 $\pm$ 0.20 &    \bf{2.97 $\pm$ 0.19}    & 3.21 $\pm$ 0.19      &    & -2.66 $\pm$ 0.06          & \bf{-2.46 $\pm$ 0.06}     & -2.58 $\pm$ 0.05 \\
Concrete & 6.16 $\pm$ 0.13 &    \bf{5.23 $\pm$ 0.12}    & 5.58 $\pm$ 0.19      &    & -3.25 $\pm$ 0.02          & \bf{-3.04 $\pm$ 0.02}     & -3.13 $\pm$ 0.03 \\
Energy   & 0.97 $\pm$ 0.09 &    1.66 $\pm$ 0.04         & \bf{0.64 $\pm$ 0.03} &    & -1.45 $\pm$ 0.10          & -1.99 $\pm$ 0.02          & \bf{-1.12 $\pm$ 0.01} \\
Kin8nm   & 0.08 $\pm$ 0.00 &    0.10 $\pm$ 0.00         & \bf{0.08 $\pm$ 0.00} &    & \bf{1.07 $\pm$ 0.00}      & 0.95 $\pm$ 0.01           & 1.06 $\pm$ 0.00 \\
Naval    & 0.00 $\pm$ 0.00 &    0.01 $\pm$ 0.00         & \bf{0.00 $\pm$ 0.00} &    & 4.61 $\pm$ 0.01           & 3.80 $\pm$ 0.01           & \bf{4.76 $\pm$ 0.00} \\
Power    & 4.21 $\pm$ 0.03 &    \bf{4.02 $\pm$ 0.04}    & 4.16 $\pm$ 0.04      &    & -2.86 $\pm$ 0.01          & \bf{-2.80 $\pm$ 0.01}     & -2.84 $\pm$ 0.01 \\
Wine     & 0.64 $\pm$ 0.01 &    \bf{0.62 $\pm$ 0.01}    & 0.65 $\pm$ 0.01      &    & -0.97 $\pm$ 0.01          & \bf{-0.93 $\pm$ 0.01}     & -0.97 $\pm$ 0.01 \\
Yacht    & 1.13 $\pm$ 0.06 &    1.11 $\pm$ 0.09         & \bf{1.08 $\pm$ 0.06} &    & -1.56 $\pm$ 0.02          & \bf{-1.55 $\pm$ 0.03}     & -1.88 $\pm$ 0.01 \\
\hline
\end{tabular}
\end{table}

Next, we present results on the UCI regression datasets which are common benchmarks for Bayesian neural networks \cite{hernandez15pbp}.
We repeat the setup\footnote{We use the data splits available at \url{https://github.com/yaringal/DropoutUncertaintyExps}} used in Gal and Ghahramani~\cite{yarin16dropout}.
Following their work, we use neural networks with one hidden layer with 50 hidden units and ReLU activation functions.
We compare SLANG with $L=1$ to the Bayes By Backprop (BBB) method \cite{blundell2015weight} and the Bayesian Dropout method of \cite{yarin16dropout}.
For the 5 smallest datasets, we used a mini-batch size of 10 and 4 Monte-Carlo samples during training.
For the 3 larger datasets, we used a mini-batch size of 100 and 2 Monte-Carlo samples during training.
More details are given in Appendix \ref{appendix:uci}.
We report test RMSE and test log-likelihood in Table \ref{table:uci}.
SLANG with just one rank outperforms BBB on 7 out of 8 datasets for RMSE and on 5 out of 8 datasets for log-likelihood.
Moreover, SLANG shows comparable performance to Dropout.

Finally, we report results for classification on MNIST.
We train a BNN with two hidden layers of 400 hidden units each.
The training set consists of 50,000 examples and the remaining 10,000 are used as a validation set.
The test set is a separate set which consists of 10,000 examples.
We use SLANG with $L = 1,2,4,8,16,32$.
For each value of $L$, we choose the prior precision and learning rate based on performance on the validation set.
Further details can be found in Appendix \ref{appendix:mnist}.
The test accuracies are reported in Table \ref{table:mnist} and compared to the results obtained in \cite{blundell2015weight} by using BBB.
For SLANG, a good performance can be obtained for a moderate $L$.
Note that there might be small differences between our experimental setup and the one used in \cite{blundell2015weight} since BBB implementation is not publicly available.
Therefore, the results might not be directly comparable.
Nevertheless, SLANG appears to perform well compared to BBB.

\begin{table}[t]
\centering
\caption{Comparison of SLANG on the MNIST dataset. We use a two layer neural network with 400 units each. SLANG obtains good performances for moderate values of $L$.}
\begin{tabular}{lcccccccc}
	  & & \hspace{0.1cm} & \multicolumn{6}{c}{SLANG} \\
	 \cmidrule(lr){4-9}
	 & \textbf{BBB} &  &\textbf{L = 1 }& \textbf{L = 2 }& \textbf{L = 4 }& \textbf{L = 8 }& \textbf{L = 16} & \textbf{L = 32}\\
	\hline
    Test Error & 1.82\% & & 2.00\% & 1.95\% & 1.81\%  & 1.92\%  & 1.77\% & {\bf 1.73\%} \\
\end{tabular}
\label{table:mnist}
\end{table}


\section{Conclusion}

We consider the challenging problem of uncertainty estimation in large deep models.
For such problems, it is infeasible to form a Gaussian approximation to the posterior distribution.
We address this issue by estimating a Gaussian approximation that uses a covariance with low-rank plus diagonal structure.
We proposed an approximate natural-gradient algorithm to estimate the structured covariance matrix.
Our method, called SLANG, relies only on the back-propagated gradients to estimate the covariance structure, which is a desirable feature when working with deep models.
Empirical results strongly suggest that the accuracy of our method is better than those obtained by using mean-field methods.
Moreover, we observe that, unlike mean-field methods, our method does not drastically shrink the marginal variances.
Experiments also show that SLANG is highly flexible and that its accuracy can be improved by increasing the rank of the covariance's low-rank component.
Finally, our method converges faster than the mean-field methods and can sometimes converge as fast as VI methods that use a full-Gaussian approximation.

The experiments presented in this paper are restricted to feed-forward neural networks.
This is partly because existing deep-learning software packages do not support individual gradient computations.
Individual gradients, which are required in line 6 of Algorithm \ref{alg:slang}, must be manually implemented for other types of architectures.
Further work is therefore necessary to establish the usefulness of our method on other types of network architectures.

SLANG is based on a natural-gradient method that employs the empirical Fisher approximation \cite{khan2017vprop}.
Our empirical results suggest that this approximation is reasonably accurate.
However, it is not clear if this is always the case.
It is important to investigate this issue to gain better understanding of the effect of the approximation, both theoretically and empirically.

During this work, we also found that comparing the quality of covariance approximations is a nontrivial task for deep neural networks.
We believe that existing benchmarks are not sufficient to establish the quality of an approximate Bayesian inference method for deep models.
An interesting and useful area of further research is the development of good benchmarks that better reflect the quality of posterior approximations.
This will facilitate the design of better inference algorithms.

\subsection*{Acknowledgements}
We thank the anonymous reviewers for their helpful feedback.
We greatly appreciate useful discussions with Shun-ichi Amari (RIKEN),
Rio Yokota (Tokyo Institute of Technology), Kazuki Oosawa (Tokyo Institute of Technology), Wu Lin (University of British Columbia), and Voot Tangkaratt (RIKEN).
We are also thankful for the RAIDEN computing system and its support team at the RIKEN Center for Advanced Intelligence Project, which we used extensively for our experiments.

\bibliographystyle{plain}
{\small \bibliography{main}}

\newpage
\appendix

\section{Derivation of the VOGN Update}
\label{app:vogn_deriv}
We can derive the VOGN update by using the Variational Online Newton (VON) method derived in Appendix D of \cite{khan2017vprop}. The VON updates are given as follows:
\begin{align}
\vmu_{t+1} &= \vmu_{t} - \beta_t \,\, \vSigma_{t+1}  \sqr{  \hat{\vg}(\vtheta_t) + \lambda\vmu_t}, \label{eq:Van_mean_1}\\
\vSigma_{t+1}^{-1} &= (1-\beta_t) \vSigma_t^{-1} + \,\, \beta_t \,\, \sqr{ \hat{\vH}(\vtheta_t) + \lambda\vI}, \label{eq:Van_var_1}
\end{align}
where $t$ is the iteration number, $\beta_t>0$ is the learning rate, $\vtheta_t\sim \gauss(\vtheta|\vmu_t, \vSigma_t)$, and $\hat{\vg}(\vtheta_t)$ and $\hat{\vH}(\vtheta_t)$ are the stochastic gradient and Hessian, defined respectively as follows:
\begin{align}
\hat{\vg}(\vtheta_t) &:= - \frac{N}{M} \sum_{i\in\minibatch} \vg_i(\vtheta_t),\\
\hat{\vH}(\vtheta_t) &:= - \frac{N}{M}\sum_{i\in\minibatch} \nabla_{\theta\theta}^2 \log p(\data_i|\vtheta_t).
\end{align}
$\vg_i(\vtheta_t) := \nabla_\theta \log p(\data_i|\vtheta_t)$ is the back-propagated gradient obtained on the $i$'th data example, and $\minibatch$ is a minibatch of $M$ data examples.

Dealing with Hessians can be difficult because they may not always be positive-definite and may produce invalid Gaussian approximations. Following \cite{khan2017vprop}, we approximate the Hessian by the Empirical Fisher (EF) matrix:
\begin{align}
   \textrm{EF: }\quad \hat{\vH}(\vtheta) \approx \hat{\vG}(\vtheta) := \frac{N}{M}\sum_{i\in\minibatch} \vg_i(\vtheta) \vg_i(\vtheta)^\top.
\label{eq:ggn}
\end{align}
This is also known as the Generalized Gauss-Newton approximation.
Using this approximation in \eqref{eq:Van_var_1} gives us the VOGN update of \eqref{eq:von}.

The VON update is an exact natural-gradient method and uses a single learning rate $\beta$. The VOGN update, on the other hand, is an approximate natural-gradient method because it uses the EF approximation. Due to this approximation, a single learning rate might not give good results and it can be sensible to use different learning rates for the $\vmu$ and $\vSigma$ updates. In \eqref{eq:von}, we therefore use a different learning rate for $\vmu$ (denoted by $\alpha$).


\section{An Alternative Low-Rank Update}
\label{appendix:old_slang}

We tried an alternative approach to learn the low-rank plus diagonal covariance approximation.
We call this method SLANG-OnlineEig. It forms the low-rank term in the precision approximation from an online estimate of the $L$ leading eigenvectors of $\hat{\vG}(\vtheta)$.
We now describe this procedure in detail.
Experimental results are presented and comparisons are made with SLANG.

\subsection{Approximating Natural Gradients by Online Estimation of the Eigendecomposition (SLANG-OnlineEig)}

The following eigenvalue decomposition forms the basis of SLANG-OnlineEig:
\begin{align}
   \hat{\vG}(\vtheta) \approx \vQ_{1:L} \vLambda_{1:L} \vQ_{1:L}^\top = \vQ_{1:L} \vLambda_{1:L}^{1/2} (\vQ_{1:L} \vLambda_{1:L}^{1/2})^\top.
\end{align}
We emphasize that SLANG-OnlineEig involves the decomposition of $N \hat{\vG}(\vtheta)$, rather than the updated matrix $(1-\beta)\vU_{t}\vU_{t}^\top + \beta N \hat{\vG}(\vtheta)$ as in SLANG.
This is cheaper by a factor of $O(DL^2)$, which is a marginal difference as $L \ll D$ in most applications.
To mimic the update of $\vSigma^{-1}$ in \eqref{eq:von}, we use the following ``moving-average`` update for $\vU$:
\begin{align} \label{eq:old_slang_low_rank_update}
   \vU_{t+1} &= (1-\beta_t) \vU_t + \beta_t \vQ_{1:L} \vLambda_{1:L}^{1/2},
\end{align}
where $\beta_t \in [0,1]$ is a scalar learning rate.
$\vU_{t+1}$ is a thus an online estimate of weighted eigenvectors of the EF.

Similar to SLANG, the diagonal $\vD$ is updated to capture the curvature information lost in the projection of $\hat{\vG}(\vtheta)$ to $\vQ_{1:L} \vLambda_{1:L} \vQ_{1:L}^\top$, i.e., the remaining $M-L$ eigenvectors:
\begin{align}
   \vD_{t+1} &= (1-\delta_t) \vD_t + \delta_t \sqr{\diag(\hat{\vG}(\vtheta_t))  - \diag(\vQ_{1:L} \vLambda_{1:L} \vQ_{1:L}^\top + \lambda\vI }. \label{eq:digcorr}
\end{align}
The updated covariance is then given by
\begin{align}
    \hat{\vSigma}_{t+1}^{-1} := \vU_{t+1}\vU_{t+1}^\top + \vD_{t+1}.
\end{align}
The final step of SLANG-OnlineEig is identical to equation \eqref{eq:slang_final_update}:
\begin{align}
   \textrm{SLANG-Online-Eig: } \quad \vmu_{t+1} &= \vmu_t - \alpha_t \hat{\vSigma}_{t+1} \sqr{ \hat{\vg}(\vtheta_t) + \lambda \vmu_t}.
\end{align}

SLANG-OnlineEig is amenable to the same algorithmic tools as SLANG, which were discussed in \ref{sec:alg_details}.
Goodfellow's trick can be used to compute the Jacobian needed for the EF and algorithms \ref{alg:fast-inverse} and \ref{alg:fast-sample} are available for fast covariance-vector products and fast sampling, respectively.
The overall computational complexity is $O(DL^2 + DLM)$ and memory cost is $O(DL + DM)$.

\subsection{Comparison with SLANG}

SLANG-OnlineEig has several promising properties.
It has slightly better computational complexity than SLANG and its update closely resembles the natural gradient update in \eqref{eq:von}.
However, update \eqref{eq:old_slang_low_rank_update} involves the product approximation,
\begin{align*}
    \vU_{t+1}\vU_{t+1}^\top &= \left((1-\beta_t) \vU_t + \beta_t \vQ_{1:L} \vLambda_{1:L}^{1/2}\right)\left((1-\beta_t) \vU_t + \beta_t \vQ_{1:L} \vLambda_{1:L}^{1/2}\right)^\top\\
    & \approx (1-\beta_t) \vU_t\vU_t^\top + \beta_t \vQ_{1:L} \vLambda_{1:L}^{1/2}\left(\vQ_{1:L} \vLambda_{1:L}^{1/2}\right)^\top,
\end{align*}
where the second line is the true product of interest.
The update assumes that the online estimate of the factors of the eigendecomposition well-approximates the online estimate of the eigendecomposition itself.

SLANG does not require the product approximation for efficient covariance learning.
Instead, $\vU_{t}\vU_{t}^\top$ is updated exactly before the projection into the space of rank-$L$ matrices.
This is why SLANG with $L = D$ reduces to \ref{eq:von} using the EF approximation.
SLANG-OnlineEig does not posses this property because of the product approximation.
It also does not necessarily match precision diagonals with the $L = D$ update, as SLANG does for all $L < D$.

A final issue is that SLANG-OnlineEig also requires matching stochastic eigenvector estimates to their corresponding online estimates in $\vU_{t+1}$.
This may introduce additional approximation error when $\hat{\vG}(\vtheta)$ is highly stochastic.

\subsection{Experimental Results for SLANG-OnlineEig}

\begin{table}[]
\centering
\caption{
Comparison of SLANG and SLANG-OnlineEig for logistic regression. SLANG obtains as good or better loss under every metric for each dataset.
Additionally, the quality of posterior approximations computed by SLANG improves while SLANG-OnlineEig approximations sometimes degrade as $L$ is increased.
The best result for each method is in bold.}
\begin{tabular}{llccccccc}
                                                                         &  & \multicolumn{3}{c}{SLANG-OnlineEig} & \hspace{0.1cm} & \multicolumn{3}{c}{SLANG} \\ \cmidrule(lr){3-5} \cmidrule(lr){7-9}
                                                                         & Metrics                        & \textbf{L = 1} & \textbf{L = 2} & \textbf{L = 5} & & \textbf{L = 1} & \textbf{L = 2} & \textbf{L = 5} \\
																		\cmidrule(lr){1-9}
\multicolumn{1}{c}{\multirow{3}{*}{Australian}}                          & ELBO                           & 0.580   & \textbf{0.578}   & 0.652 &  & 0.574  & 0.574  & \textbf{0.569}    \\
\multicolumn{1}{c}{}                                                     & NLL                            & 0.344   & \textbf{0.343}   & 0.347 &  & 0.342  & 0.342  & \textbf{0.339}    \\
\multicolumn{1}{c}{}                                                     & KL {\small ($\times 10^4$)}    & 0.057   & 0.032   & \textbf{0.012} &  & 0.033  & 0.031  & \textbf{0.008}    \\
\cmidrule(lr){1-9}
\multirow{3}{*}{\begin{tabular}[c]{@{}l@{}}Breast\\ Cancer\end{tabular}} & ELBO                           & \textbf{0.114}   & 0.116   & 0.123 &  & 0.112  & 0.111  & \textbf{0.111}    \\
                                                                         & NLL                            & \textbf{0.092}   & 0.093   & 0.093 &  & 0.092  & 0.092  & \textbf{0.092}    \\
                                                                         & KL {\small ($\times 10^0$)}    & \textbf{1.544}   & 2.128   & 4.402 &  & 0.911  & 0.756  & \textbf{0.842}    \\
\cmidrule(lr){1-9}
\multirow{3}{*}{a1a}                                                     & ELBO                           & 0.380   & \textbf{0.380}   & 0.383 &  & 0.377  & 0.376  & \textbf{0.374}    \\
                                                                         & NLL                            & 0.339   & 0.339   & \textbf{0.339} &  & 0.339  & 0.339  & \textbf{0.339}    \\
                                                                         & KL {\small ($\times 10^2$)}    & 0.351   & 0.293   & \textbf{0.253} &  & 0.305  & 0.249  & \textbf{0.173}    \\
\cmidrule(lr){1-9}
\multirow{3}{*}{USPS 3vs5}                                               & ELBO                           & 0.210   & \textbf{0.208}   & 0.210 &  & 0.210  & 0.206  & \textbf{0.198}    \\
                                                                         & NLL                            & 0.133   & \textbf{0.132}   & 0.133 &  & 0.132  & 0.132  & \textbf{0.132}    \\
                                                                         & KL {\small ($\times 10^1$)}    & 1.497   & \textbf{1.353}   & 1.432 &  & 1.492  & 1.246  & \textbf{0.755}    \\
\cmidrule(lr){1-9}
\end{tabular}
\label{tbl:comparison_with_old_slang}
\end{table}

Table \ref{tbl:comparison_with_old_slang} compares SLANG and SLANG-OnlineEig for logistic regression on the Australian, Breast Cancer, USPS 3vs5, and a1a datasets from LIBSVM.
The results presented for SLANG are identical to those in Table \ref{tbl:logRegExperiments}.
SLANG always matches or beats the best results for SLANG-OnlineEig.
Furthermore, as $L$ is increased, the quality of posterior approximations computed by SLANG improves while SLANG-OnlineEig approximations sometimes degrade.
We speculate that this is due to the product approximation.

\begin{figure*}[ttt!]
\centering
{\includegraphics[width=3.9in]{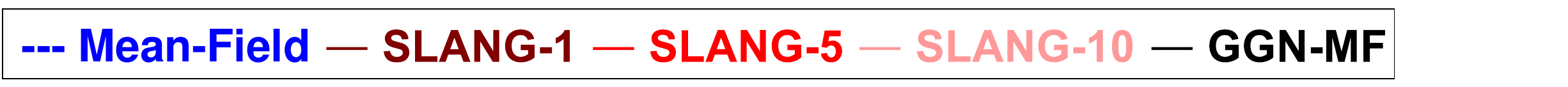} \label{fig:logreg_neg_avg_elbo}}\\
\subfigure[]%
{\includegraphics[width=1.8in]{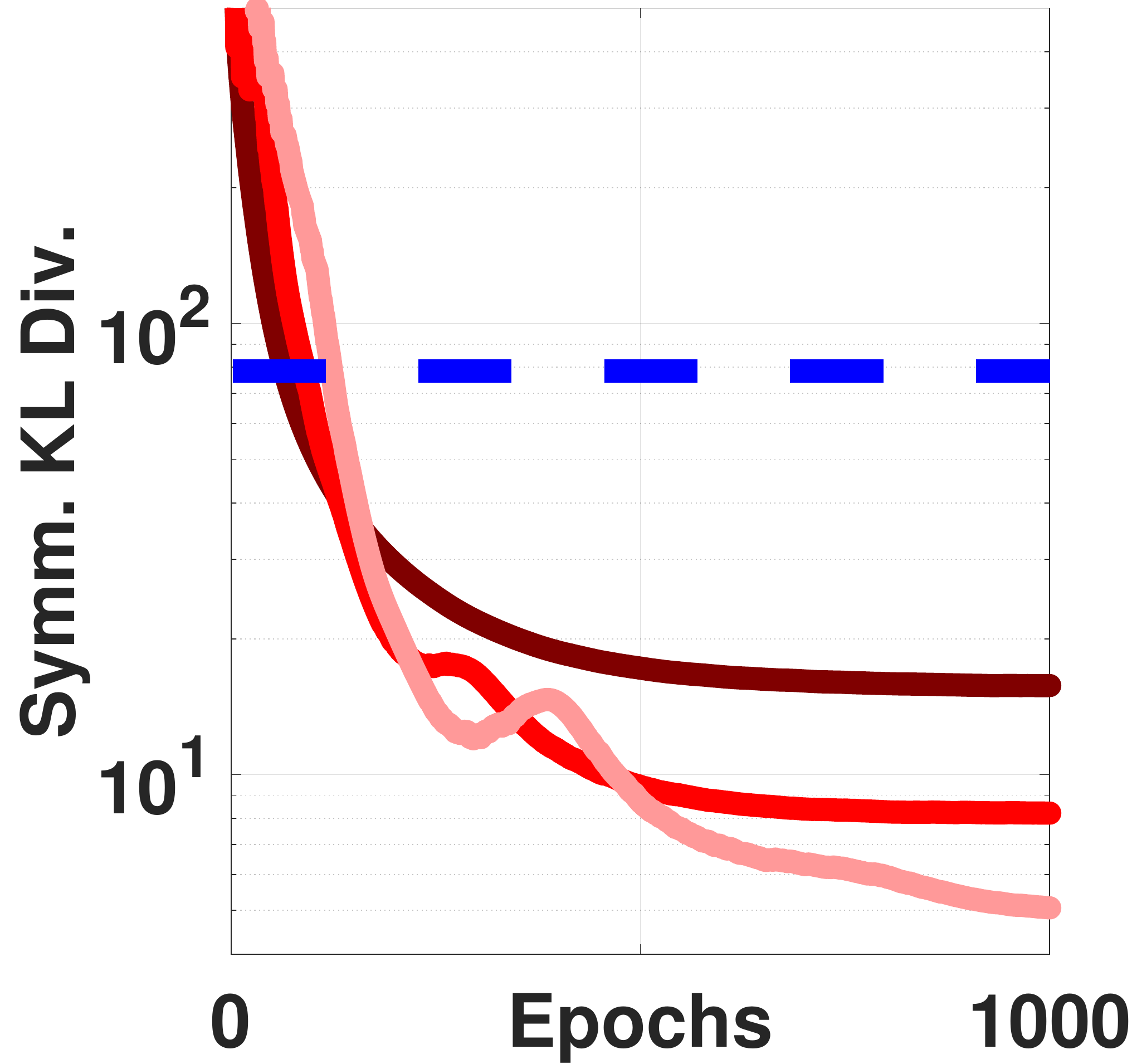} \label{fig:logreg_kl_with_full}}
\hfill
\subfigure[]%
{\includegraphics[width=1.65in]{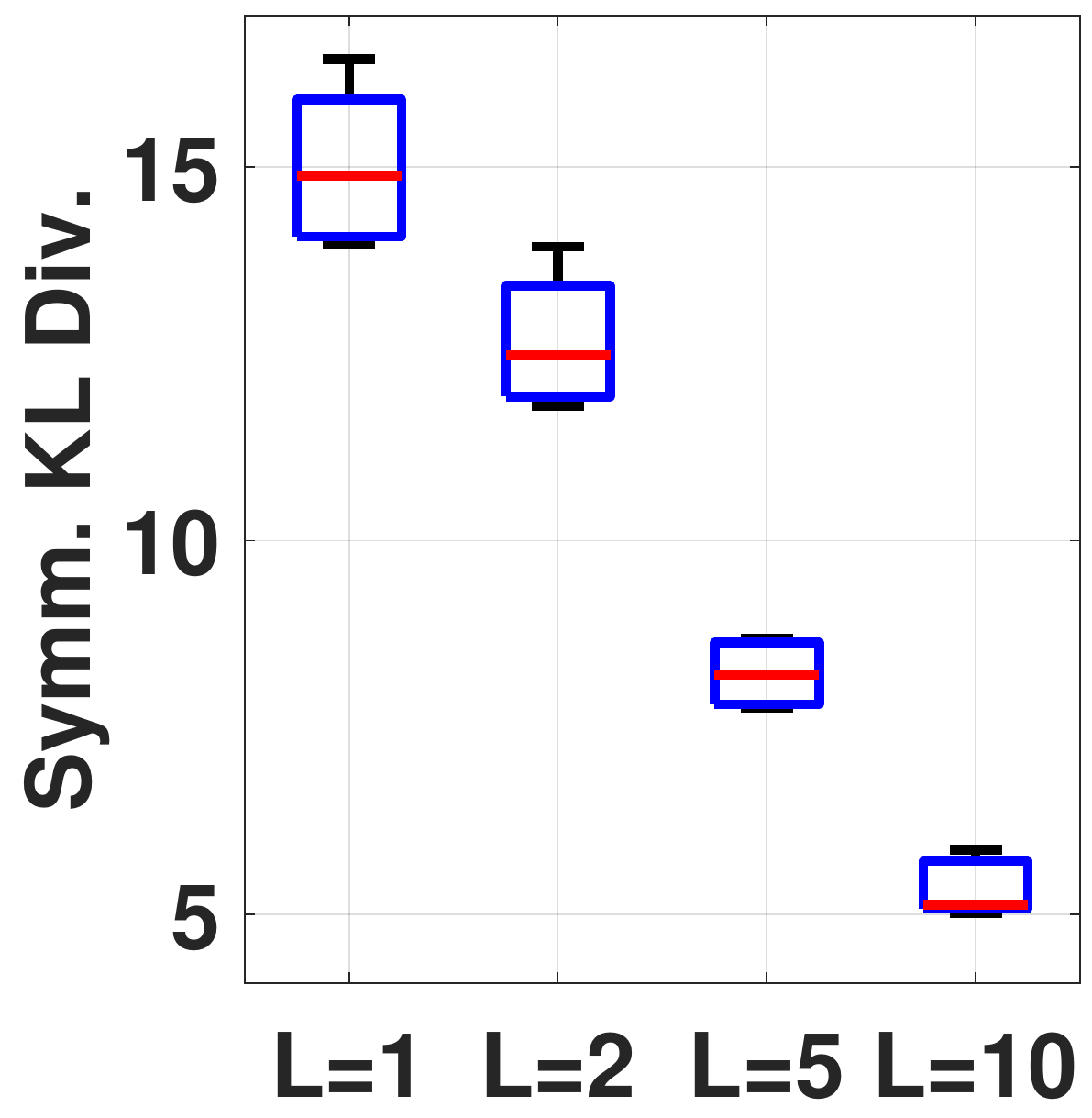} \label{fig:logreg_kl_boxplot}}
\hfill
\subfigure[]%
{\includegraphics[width=1.7in]{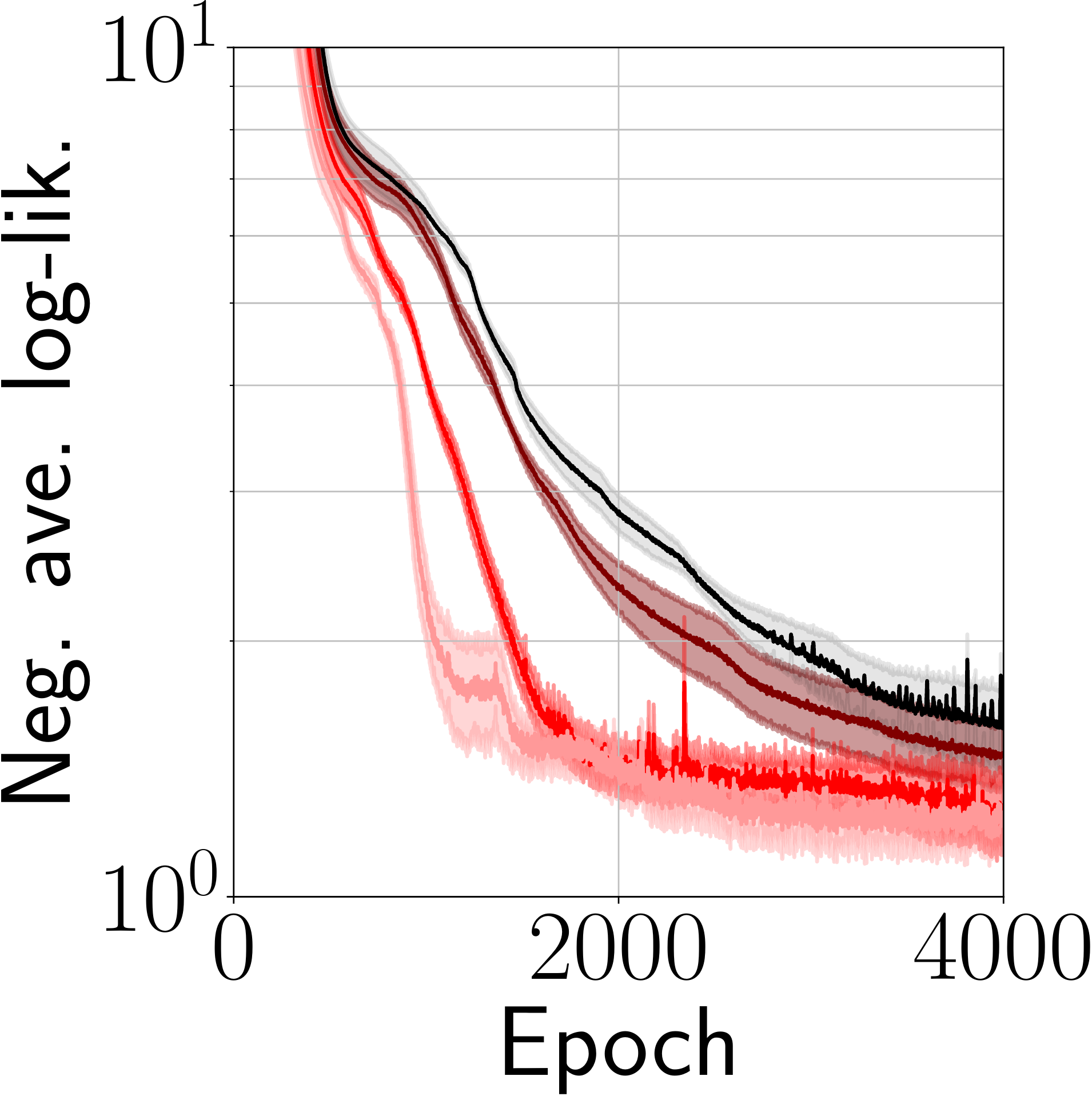} \label{fig:bnn_log_lik}}
\caption{(Left) This figure shows the convergence behavior of SLANG-OnlineEig on logistic regression for USPS.
We plot the KL divergence between Full-Gaussian and the approximate posterior.
The higher rank approximations give better results and all of them beat mean-field.
(Middle) This figure summarizes the results of 5 runs, showing that as we increase $L$ the approximation gets better.
(Right) We show convergence of a Bayesian neural network on the Energy dataset.
We see that better structured approximation leads to faster convergence.
}
\label{fig:online-eig-convergence}
\end{figure*}

Figure \ref{fig:online-eig-convergence} presents results on the convergence of SLANG-OnlineEig for logistic regression and regression with a Bayesian neural network.

Table \ref{table:uci_onlineeig} shows regression on the UCI datasets using Bayesian neural networks.
The setup for this experiment was similar to the experiment in Section \ref{sec:uci}, except that the learning rates were fixed to $\alpha = 0.01$ and $\beta = (0.9, 0.999)$ for all datasets, both for SLANG-OnlineEig and for the Adam optimizer used for BBB.
Moreover, the search spaces for the Bayesian optimization were fixed (using a normalized scale for the noise precision) and not adjusted to the individual datasets.
Finally, BBB used 40 MC samples for the 5 smallest datasets and 20 MC samples for the 3 largest datasets in this experiment.

\begin{table}[t]
\setlength{\tabcolsep}{4pt}
\centering
\caption{Predictive performance on UCI datasets using Bayesian neural networks where SLANG-OnlineEig beats BBB and performs comparably to Dropout.}
\label{table:uci_onlineeig}
\begin{tabular}{l c c c c c c c}
& \multicolumn{3}{c}{Test RMSE} & \hspace{0.1cm} & \multicolumn{3}{c}{Test log-likelihood} \\
\cmidrule{2-4}
\cmidrule{6-8}
\bf{Dataset} & \bf{BBB} & \bf{Dropout} & \bf{OnlineEig} & & \bf{BBB} & \bf{Dropout} & \bf{OnlineEig} \\
\cmidrule{1-8}
Boston   & 4.04 $\pm$ 0.28 & \bf{2.97 $\pm$ 0.19} & 3.17 $\pm$ 0.17 & & -2.75 $\pm$ 0.07 & \bf{-2.46 $\pm$ 0.06} & -2.61 $\pm$ 0.06 \\
Concrete & 6.16 $\pm$ 0.14 & \bf{5.23 $\pm$ 0.12} & 5.79 $\pm$ 0.13 & & -3.22 $\pm$ 0.02 & \bf{-3.04 $\pm$ 0.02} & -3.19 $\pm$ 0.02 \\
Energy   & 0.86 $\pm$ 0.04 & 1.66 $\pm$ 0.04 & \bf{0.59 $\pm$ 0.01} & & -1.20 $\pm$ 0.05 & -1.99 $\pm$ 0.02 & \bf{-1.05 $\pm$ 0.01} \\
Kin8nm   & 0.09 $\pm$ 0.00 & 0.10 $\pm$ 0.00 & \bf{0.08 $\pm$ 0.00} & & 0.97 $\pm$ 0.01 & 0.95 $\pm$ 0.01 & \bf{1.13 $\pm$ 0.00} \\
Naval    & \bf{0.00 $\pm$ 0.00} & 0.01 $\pm$ 0.00 & \bf{0.00 $\pm$ 0.00} & & \bf{5.34 $\pm$ 0.07} & 3.80 $\pm$ 0.01 & 5.09 $\pm$ 0.08 \\
Power    & 4.28 $\pm$ 0.03 & \bf{4.02 $\pm$ 0.04} & 4.09 $\pm$ 0.04 & & -2.87 $\pm$ 0.01 & \bf{-2.80 $\pm$ 0.01} & -2.83 $\pm$ 0.01 \\
Wine     & 0.66 $\pm$ 0.01 & \bf{0.62 $\pm$ 0.01} & 0.64 $\pm$ 0.01 & & -0.99 $\pm$ 0.01 & \bf{-0.93 $\pm$ 0.01} & -0.98 $\pm$ 0.01 \\
Yacht    & 1.07 $\pm$ 0.08 & 1.11 $\pm$ 0.09 & \bf{0.72 $\pm$ 0.04} & & -1.45 $\pm$ 0.03 & -1.55 $\pm$ 0.03 & \bf{-1.41 $\pm$ 0.01} \\
\cmidrule{1-8}
\end{tabular}
\end{table}


\section{Additional Algorithmic Details for SLANG}
\label{apx:algorthmic_details}
The following section gives more detail about the individual components of the algorithm required to leverage the low-rank plus diagonal structure for computational efficiency.
In general, we obtain efficient algorithms by operating only on $D \times L$ or $L \times L$ matrices.
This avoids the $O(D^2)$ storage cost and the $O(D^3)$ computational cost of working in the $D \times D$ space.

\subsection{Fast Computation of Individual Gradients}
\label{apx:backprop-goodfellow}
Most deep-learning automatic differentiation packages,
such as PyTorch \cite{paszke2017automatic} and TensorFlow \cite{abadi2016tensorflow},
are optimized to return the \emph{overall} gradient of a minibatch, not \emph{individual} gradients for each example passed through the network.
It is true that the naïve option of doing a forward and backward pass for each example has a similar computational complexity as a fully parallel version.
However, in practice most of the computations involved can be either reused across examples, or sped up drastically by batching them.
Implementations that use matrix-matrix multiplications instead of repeatedly doing matrix-vector operations are far more efficient on GPUs.

Ian Goodfellow's note \cite{goodfellow2015efficient} outlines a method for efficiently computing per-example gradients.
It suggests saving the neuron activations and the linear combinations of activations computed during the minibatch's forward pass through the neural network.
These values are then used in a manual implementation of the per-example gradients, which avoids the summation defined by the cost function.
While much more efficient than the sequential approach, Goodfellow's approach requires more implementation effort;
a separate implementation is required to handle each type of layer used.
This is partly why the experiments presented here are limited to standard Multi-Layer Perceptrons.
We hope to improve upon this in future implementations.

\subsection{Fast Top-$L$ Eigendecomposition}
\label{apx:eigendecomposition}
The goal is to get the top-$L$ eigenvalues and eigenvectors of the matrix
$(1-\beta) \vU_t\vU_t^\top + \beta \hat{\vG}(\vtheta_t)$ defined in \eqref{eq:deriv_update_2}.
Since we do not want to compute the $D \times D$ matrix explicitly,
we use the low-rank structure of the update matrix to compute matrix-vector or matrix-matrix products with computations in $O(DL + DM)$.
This can be seen by rewriting the matrix as follows:
\begin{equation}
	(1-\beta) \vU_t\vU_t^\top + \beta \hat{\vG}(\vtheta_t)
	= (1-\beta) \sum_{l=1}^L \vu_t^{(l)}\vu_t^{(l)}{}^\top + \beta \frac{N}{M} \sum_{i \in \minibatch} \vg_i(\vtheta_t) \vg_i(\vtheta_t)^\top.
\end{equation}
These products can be used to compute eigendecomposition efficiently by using a randomized algorithm.

The main idea behind the randomized eigendecomposition is to project a matrix $\vA$ onto a randomly selected subspace by sampling $K$ vectors $\vepsilon_k \in \mathbb{R}^D$, each entry being selected uniformly at random, and computing $\vA_K = \vA [\vepsilon_1, ..., \vepsilon_K]$, where $K$ is larger than $L$.
A traditional eigendecomposition can then be performed on the $D \times K$ matrix $\vA_K$ to recover the top $L$ eigenvectors, with $K$ acting as a precision-computation tradeoff parameter.
More details on randomized eigenvalue methods can be found in \cite{halko2011finding}.

Our implementation of this procedure follows Facebook's Fast Randomized SVD\footnotemark closely;
starting with a random matrix, multiplies it by $\vA$ and applies a QR decomposition on the result.
This process is repeated on the resulting matrix for a few iterations to improve stability, similarly to the Lanczos iterations.
As all operations are done on the smaller $D \times K$ matrix, using $K = L + 2$ as recommended in \textsuperscript{\ref{ftn:1}}), the computational cost of the QR decomposition and eigendecomposition are in $O(DL^2)$, leading to an $O(DL^2 + DM)$ algorithm overall.
\footnotetext{
\label{ftn:1}
\href{https://github.com/facebook/fbpca}{https://github.com/facebook/fbpca},
\href{https://research.fb.com/fast-randomized-svd/}{https://research.fb.com/fast-randomized-svd/}
}

\subsection{Fast multiplication by inverse of low-rank + diagonal}
\label{apx:fast-inverse}
To implement the natural gradient update \eqref{eq:slang_final_update}, we need to be able to multiply an arbitrary vector by $\hat{\vSigma}$, given $\hat{\vSigma}^{-1} = \vU \vU^\top + \vD$.
Woodbury's identity can be used to do so without forming the $D \times D$ matrix and doing the costly $O(D^3)$ inversion. The identity gives
\begin{equation}
(\vD + \vU\vU^\top)^{-1} = \vD^{-1} - \vD^{-1} \vU(\vI_L + \vU^\top \vD^{-1} \vU)^{-1} \vU^\top \vD^{-1}.
\end{equation}
The only inversions remaining involve diagonal or $L \times L$ matrices.
Correct ordering of the operations allows the $O(D^2)$ storage cost to be avoided when computing the product $\big( \vU \vU^\top + \vD \big)^{-1}\vx$ and ensures that we only need to store $D \times L, L \times L$ or diagonal matrices,
\begin{equation}
	\left(\vD + \vU\vU^\top\right)^{-1}\vx
	= \left(\vD^{-1}\vx\right) -
	\vD^{-1} \left(\vU \left(\left(\vI_L + \vU^\top \vD^{-1} \vU\right)^{-1} \left( \vU^\top \left(\vD^{-1} \vx\right)\right)\right)\right),
\end{equation}
yielding a $O(DL^2)$ algorithm.

\subsection{Fast sampling}
\label{apx:fast-sample}
To generate a sample from $\mathcal{N}(\vmu, \hat{\vSigma})$, it is sufficient to generate a sample $\vepsilon \sim \mathcal{N}(0, \vI_D)$ and compute
$\vmu + \vA \vepsilon$, where $\vA\vA^\top = \hat{\vSigma}$. $\hat{\vSigma}$ can be factorized efficiently by exploiting its "low-rank plus diagonal" structure:
\begin{align}
	\hat{\vSigma}
	&= \left(\vU\vU^\top + \vD\right)^{-1},\\
	&= \left(\vD^{1/2}\left(\underbrace{\vD^{-1/2}\vU\vU^\top\vD^{-1/2}}_{\vV\vV^\top} + \vI\right)\vD^{1/2}\right)^{-1},\\
	&= \vD^{-1/2}\left(\vV\vV^\top + \vI\right)^{-1}\vD^{-1/2}.
\end{align}
Letting $\vW$ be a symmetric factor for $\vV\vV^\top + \vI$, we then have that $\vD^{-1/2}\vW^{-1}$ is a symmetric factor for $\hat{\vSigma}$.
Such a factorization can be found using the work of \cite{ambikasaran2014fast}, which showed that by taking
\begin{equation}
	\begin{aligned}
		\vA &= \text{Cholesky}(\vV^\top \vV),\\
		\vB &= \text{Cholesky}(\vV^\top \vV + \vI_L),\\
		\vC &= \vA^{-\top}(\vB - \vI_L) \vA^{-1},
	\end{aligned}
\end{equation}
$\vW = \vI_D + \vV \vC \vV^\top$ is a symmetric factorization for $\vI_D + \vV \vV^\top$.
We can then use Woodbury's Identity to avoid taking the inverse in the $D \times D$ space,
\begin{equation}
\vD^{-1/2}\left(\vI_D + \vV \vC \vV^\top\right)^{-1}\vepsilon
=
\vD^{-1/2}\left(\vI_D - \vV \left(\vC^{-1} + \vV^\top\vV\right)^{-1}\vV^\top\right)\vepsilon
\end{equation}
and careful ordering of operations, as above, leads to a $O(DL^2)$ complexity. This subroutine is implemented in Algorithm \ref{alg:fast-sample}.

\section{Diagonal Correction}
\label{:appendix_diagonal_correction}

In this section, we prove that the diagonal of the precision computed by SLANG when $L < D$ is identical to the diagonal computed when $L = D$.

Consider the precision $\hat \vSigma_{t}^{-1} = \vD_t + \vU_t \vU_t^\top$, where the diagonal and low rank components are updated by \eqref{eq:low_rank_update}, \eqref{eq:slang_diagonal_difference} and \eqref{eq:slang_diagonal_update}.
Recall that when $L=D$,
\[ \hat \vSigma_{t}^{-1} = \vSigma_{t}^{-1}, \]
where $\vSigma_{t}^{-1}$ was the precision matrix updated by \eqref{eq:von}.
Assume both methods use the same initial diagonal precision matrix and that they are updated by same sequence of EF matrices $\{\hat{\vG}(\vtheta)\}$.
Then we will show that at every iteration $t$,
\[ \text{diag} \Big[\hat \vSigma_{t}^{-1}\Big] = \text{diag}\Big[\vSigma_{t}^{-1}\Big]. \]

\textbf{Proof:}

$\hat \vSigma_{t}^{-1}$ and $\vSigma_{t}^{-1}$ are initialized as the same diagonal matrix, so the claim holds trivially at $t = 0$.\\

Assume that the claim holds at some iteration $t$. The inductive hypothesis implies
\[ \text{diag} \Big[ \vSigma_{t}^{-1} \Big] = \text{diag} \Big[\vU_t \vU_t^\top \Big] + \vD_t = \text{diag} \Big[\hat \vSigma_{t}^{-1}\Big]. \]

Applying the update \eqref{eq:von} gives the diagonal of $\vSigma_{t+1}^{-1}$ to be
\begin{align*}
\text{diag} \Big[\vSigma_{t+1}^{-1}\Big] &= (1-\beta) \ \text{diag} \Big[\vSigma_{t}^{-1}\Big]+ \beta \text{diag} \big[ \hat{\vG}(\vtheta) \big] + \beta \lambda \vI\\
&= (1-\beta) \ \text{diag} \Big[\vU_t \vU_t^\top \Big] + (1-\beta) \ \vD_t + \beta \text{diag} \big[ \hat{\vG}(\vtheta) \big] + \beta \lambda \vI
\end{align*}
The diagonal of the SLANG precision at $t+1$ is
\begin{align*}
\text{diag} \Big[\hat \vSigma_{t+1}^{-1}\Big] &= \text{diag} \Big[ \vU_{t+1} \vU_{t+1}^\top \Big] + \vD_{t+1}\\
 &=  \text{diag} \left[\vU_{t+1} \vU_{t+1}^\top\right] + (1-\beta) \vD_t + \beta \lambda \vI + \Delta_D \\
 &= (1-\beta) \ \text{diag} \Big[\vU_t \vU_t^\top \Big] + (1-\beta) \vD_t + \beta \text{diag} \big[ \hat{\vG}(\vtheta) \big] + \beta \lambda\vI,
\end{align*}
where the last line is obtained by expanding $\Delta_D$ and canceling the $\text{diag} \left[\vU_{t+1} \vU_{t+1}^\top\right]$ terms. This completes the proof. In practice, the diagonal of the update might differ because the two methods might update $\vmu$ differently. Nevertheless, the above results shows a desirable property of SLANG.


\section{Details for Experiments on Bayesian Logistic Regression}
\label{app:logreg}

We present additional results for SLANG on logistic regression and then provide algorithmic details for all logistic regression experiments.

\subsection{Additional Results}
\label{app:add_logreg}

\begin{figure*}[ttt!]
\centering
{\includegraphics[width=5.3in]{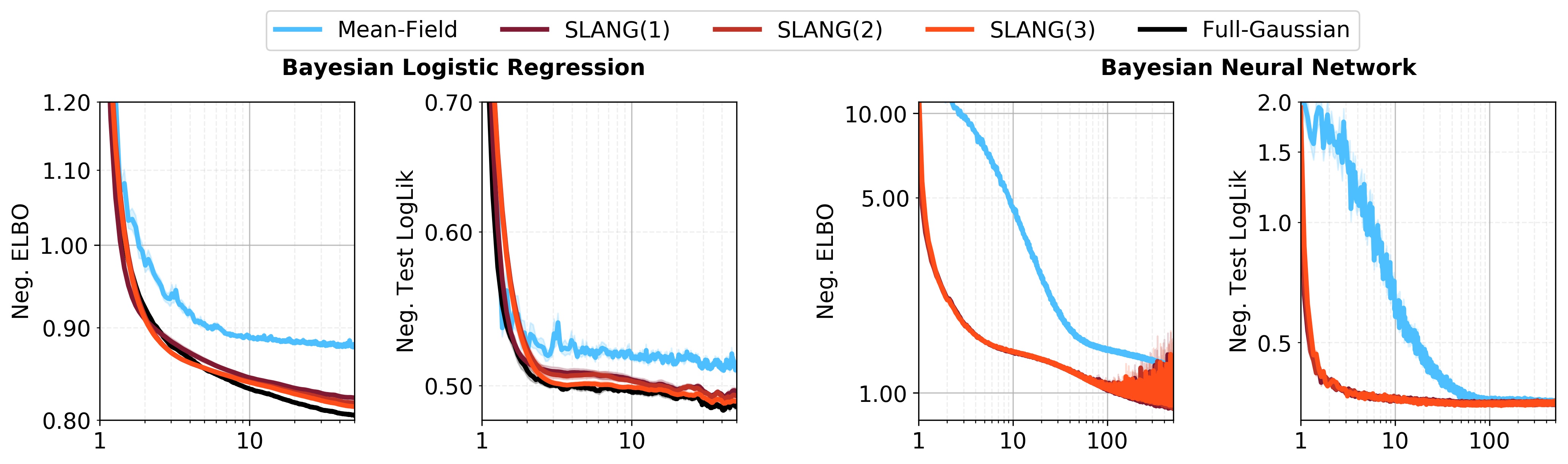}}
\caption{This figure compares the convergence behavior on Australian for two models: Bayesian logistic regression (left) and Bayesian neural networks (BNN) (right).
SLANG(1, 2, 3) refers to $L=1, 5, 10$ for logistic regression and $L=8, 16, 32$ for BNN.
The mean-field method is a natural-gradient mean-field method for logistic regression (see text) and BBB \cite{blundell2015weight} for the BNN experiment.}
\label{fig:australian_convergence}
\end{figure*}

Additional convergence results for logistic regression are provided in Figure \ref{fig:australian_convergence}, which shows the behavior of SLANG on the Australian dataset.
These results are from the same experiments as those presented in Figure \ref{fig:convergence}.

\begin{figure}[!t]
   \center
	\newcommand{\myplotscaling}{.8}
	\subfigure[Australian-scale]{\includegraphics[width=\myplotscaling\columnwidth]{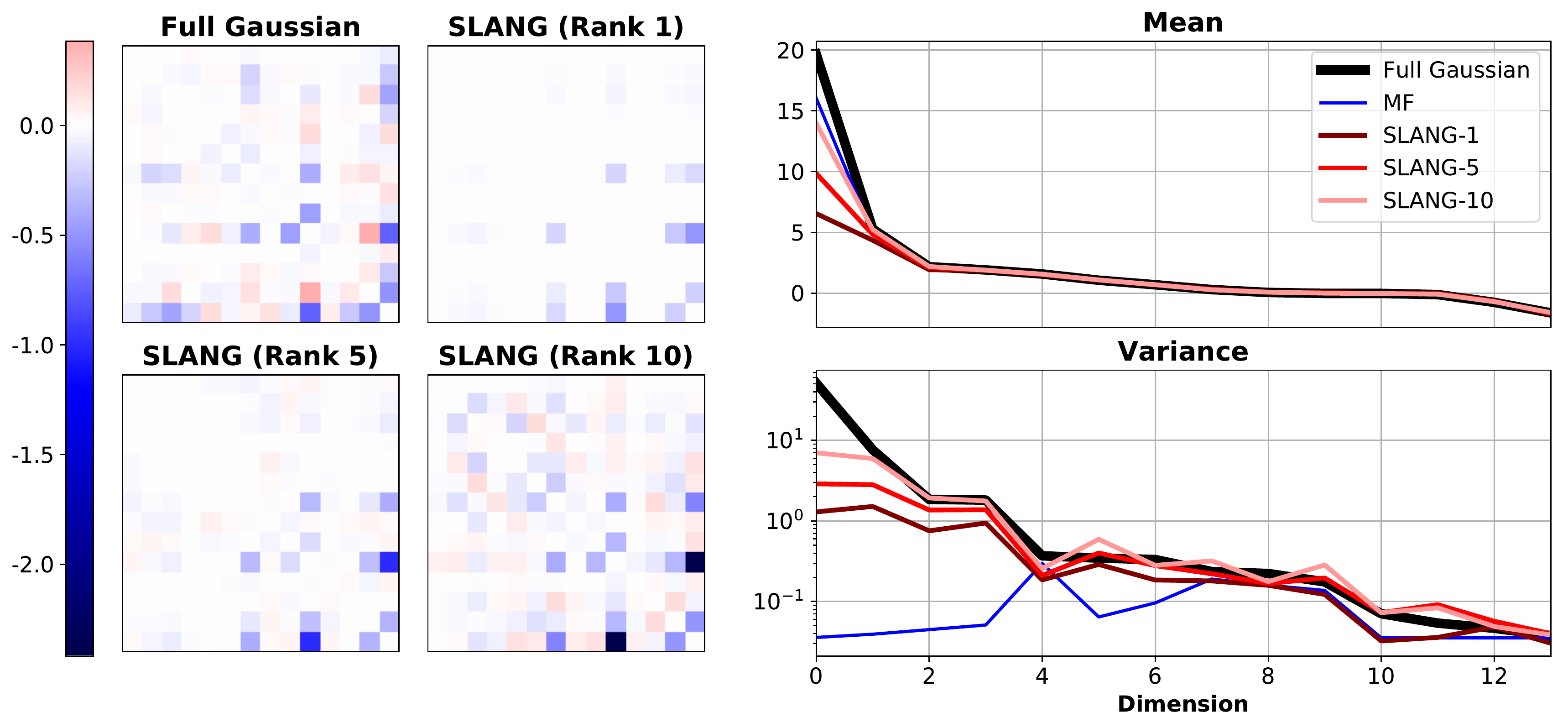}}
	\subfigure[Breast cancer]{\includegraphics[width=\myplotscaling\columnwidth]{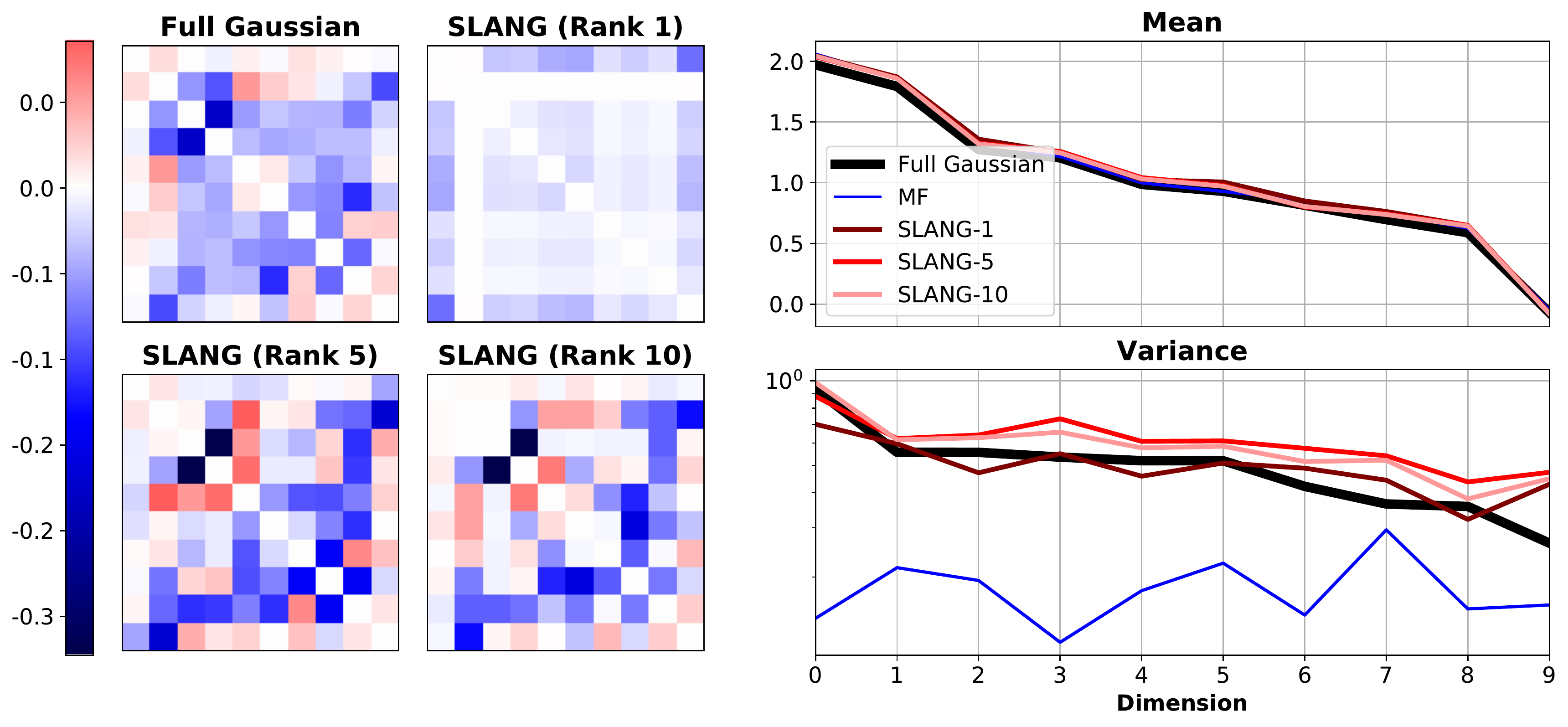}}
	\subfigure[a1a]{\includegraphics[width=\myplotscaling\columnwidth]{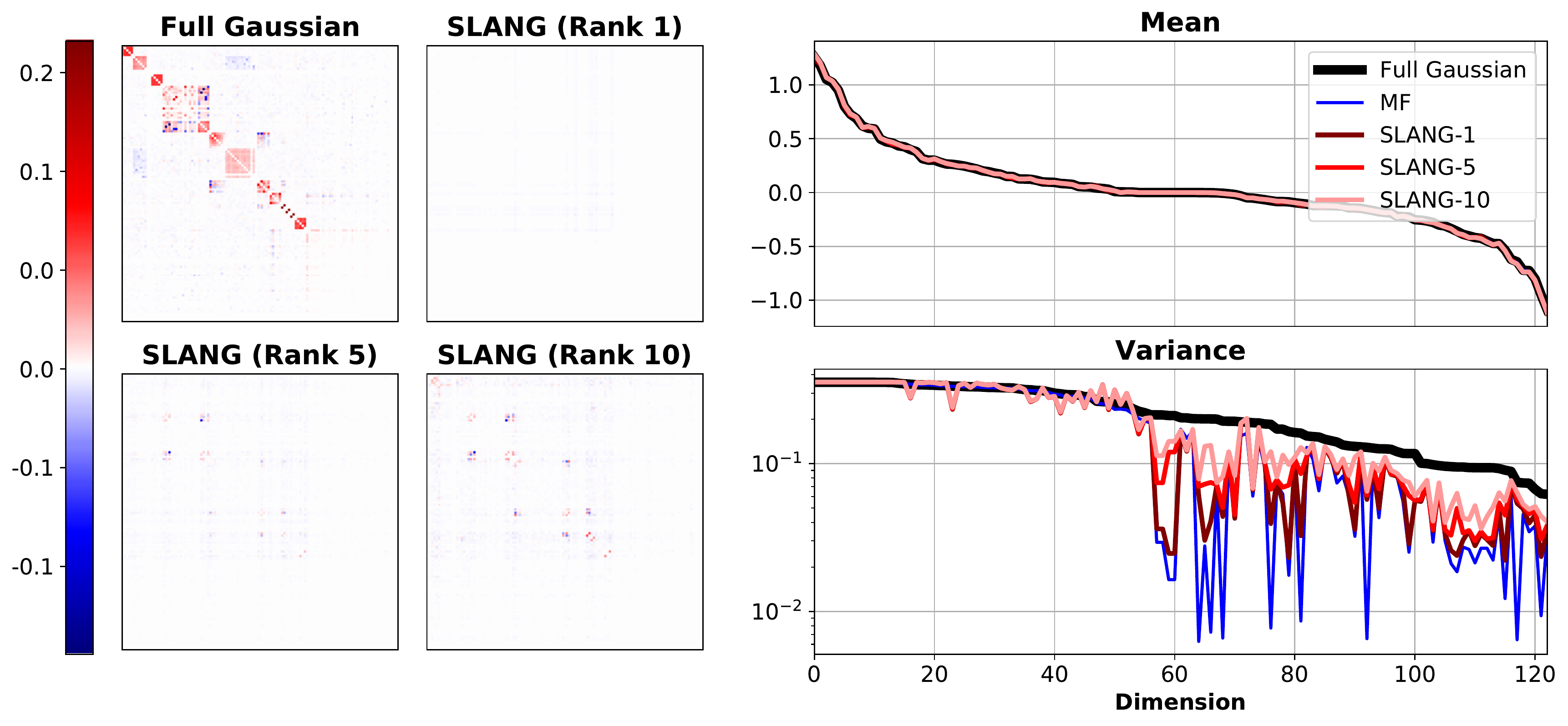}}
	\subfigure[Table of (mean, variance) for the bias term]{%
	\small \makebox[\textwidth][c]{\begin{tabular}{lrrrrr}
		{\bf Dataset}	& {\bf MF}		& {\bf SLANG-1}	& {\bf SLANG-5}	& {\bf SLANG-10}	& {\bf Full-Gaussian}	\\
		Australian 		& $19.94, 0.04$	& $10.02, 1.26$	& $13.92, 2.82$	& $18.49, 6.98$		& $24.08, 56.93$		\\
		Breast cancer	& $4.21, 0.12$	& $4.47, 1.17$	& $4.44, 1.72$	& $4.41, 1.75$		& $4.26, 1.54$			\\
		A1A 			& $-2.13, 0.01$	& $-2.04, 0.16$	& $-2.13, 0.25$	& $-2.17, 0.37$		& $-2.11, 1.37$			\\
		USPS 3 vs. 5	& $2.28, 0.03$	& $2.17, 0.67$	& $2.06, 0.95$	& $1.98, 1.38$		& $1.80, 2.09$			\\
	\end{tabular}}}

	\caption{
	Comparison of the posterior approximations of SLANG, full-Gaussian and Mean-Field (MF) methods.
	The figures on the left compare the structure of the off-diagonal covariance and the figures on the right compare the means and diagonal variances.
	While the means are closely matched for all methods, the MF approximation underestimates the variances on all three datasets.
	Note that the diagonal of the covariance is not included in the covariance plot on the left, and the bias term is only shown in the last table
	- as the off-diagonal, diagonal and bias covariances are of different magnitude, a single scale would make comparison difficult.
	}
	\label{fig:plot1-logreg}
\end{figure}

Figure \ref{fig:plot1-logreg} shows qualitative comparisons of posterior means, variances, and covariances for the Australian, Breast Cancer, and a1a datasets, similar to Figure \ref{fig:1}.
These results resemble those for USPS, where the mean-field method (MF Exact) displays "trend-reversal" for the marginal covariances when compared to the Full-Gaussian Exact method.
In comparison, SLANG gives a good approximation of the ground-truth Full-Gaussian covariance approximation for Australian and Breast Cancer.
For the a1a dataset, SLANG with L = 10 fails to learn the covariance structure and shows mixed results on the marginal variances.
We believe that this is because the dimensionality of a1a is quite large ($D = 1,605$). We expect SLANG to improve when L is sufficiently increased.

Tables \ref{tbl:logRegExperiments-baselines} and \ref{tbl:logRegExperiments-slang} are more detailed versions of Table \ref{tbl:logRegExperiments}.
The tables are split into baselines and SLANG to improve readability.
Table \ref{tbl:logRegExperiments-slang} also reports values for L = 2, which are not reported in Table \ref{tbl:logRegExperiments} due to space constraints.

\begin{landscape}
\begin{table}[]
\centering
\caption{Comparison of SLANG to many mean-field and full-Gaussian methods.
Results for mean-field and full-Gaussian methods are shown in this table, while
results for SLANG are shown in Table \ref{tbl:logRegExperiments-slang} due to space constraints.
We see that SLANG with $L=1$ shows better performance than mean-field methods.
It is also quite close to the performance of full-Gaussian method, except in a1a.
We expect SLANG to do better on a1a if we increase the rank further. }
\begin{tabular}{cccccccc}
&                                       & \multicolumn{3}{c}{Mean-Field Methods}      & \multicolumn{3}{c}{Full-Gaussian} \\ \cmidrule(lr){3-5} \cmidrule(lr){6-8}
\textbf{Datasets}                                        & \textbf{Metrics}               & \textbf{EF}                           & \textbf{Hessian}                 & \textbf{Exact}                                        & \textbf{EF}       & \textbf{Hessian}    & \textbf{Exact}    \\ \hline
\multirow{3}{*}{Australian}             & ELBO                  & 0.6139 $\pm$ 0.0059             & 0.6125  $\pm$ 0.0059 & 0.5933 $\pm$ 0.0058 & 0.5601 $\pm$ 0.0059 & 0.5583        $\pm$ 0.0059 & 0.5589 $\pm$ 0.0059 \\
                                        & NLL                   & 0.3480 $\pm$ 0.0069             & 0.3472  $\pm$ 0.0068 & 0.3413 $\pm$ 0.0072 & 0.3396 $\pm$ 0.0072 & 0.3386        $\pm$ 0.0072 & 0.3377 $\pm$ 0.0069 \\
                                        & KL {\small ($\times 10^4$)}    & 2.2398 $\pm$ 0.3459             & 2.0301  $\pm$ 0.3146 & 0.1946 $\pm$ 0.0214 & 0.0001 $\pm$ 0.0000 & 0.0000        $\pm$ 0.0000 & 0.0000 $\pm$ 0.0000 \\ \hline
\multirow{3}{*}{
\begin{tabular}[c]{@{}l@{}}Breast\\
   Cancer\end{tabular}}                 & ELBO                  & 0.1217 $\pm$ 0.0028             & 0.1208  $\pm$ 0.0028 & 0.1205 $\pm$ 0.0028 & 0.1107 $\pm$ 0.0028 & 0.1086        $\pm$ 0.0029 & 0.1087 $\pm$ 0.0029 \\
                                        & NLL                   & 0.0950 $\pm$ 0.0024             & 0.0943  $\pm$ 0.0023 & 0.0937 $\pm$ 0.0024 & 0.0920 $\pm$ 0.0023 & 0.0912        $\pm$ 0.0023 & 0.0912 $\pm$ 0.0024 \\
                                        & KL                    & 8.0188 $\pm$ 0.2540             & 9.0706  $\pm$ 0.1750 & 7.7713 $\pm$ 0.1173 & 0.6373 $\pm$ 0.0221 & 0.0017        $\pm$ 0.0003 & 0.0000 $\pm$ 0.0000 \\ \hline
\multirow{3}{*}{a1a}                    & ELBO                  & 0.3838 $\pm$ 0.0000             & 0.3833  $\pm$ 0.0000 & 0.3828 $\pm$ 0.0000 & 0.3686 $\pm$ 0.0000 & 0.3678        $\pm$ 0.0000 & 0.3679 $\pm$ 0.0000 \\
                                        & NLL                   & 0.3390 $\pm$ 0.0000             & 0.3389  $\pm$ 0.0000 & 0.3385 $\pm$ 0.0000 & 0.3386 $\pm$ 0.0000 & 0.3385        $\pm$ 0.0000 & 0.3386 $\pm$ 0.0000 \\
                                        & KL {\small ($\times 10^2$)}    & 2.5896 $\pm$ 0.0000             & 2.2082  $\pm$ 0.0000 & 1.2946 $\pm$ 0.0000 & 0.0141 $\pm$ 0.0000 & 0.0001        $\pm$ 0.0000 & 0.0000 $\pm$ 0.0000 \\ \hline
\multirow{3}{*}{
\begin{tabular}[c]{@{}l@{}}
   USPS\\ (3vs5)\end{tabular}}          & ELBO                  & 0.2679 $\pm$ 0.0029             & 0.2675  $\pm$ 0.0029 & 0.2672 $\pm$ 0.0028 & 0.1886 $\pm$ 0.0022 & 0.1860        $\pm$ 0.0022 & 0.1860 $\pm$ 0.0022 \\
                                        & NLL                   & 0.1390 $\pm$ 0.0020             & 0.1388  $\pm$ 0.0020 & 0.1383 $\pm$ 0.0020 & 0.1309 $\pm$ 0.0020 & 0.1300        $\pm$ 0.0020 & 0.1301 $\pm$ 0.0020 \\
                                        & KL {\small ($\times 10^1$)}    & 7.6836 $\pm$ 0.1485             & 7.1878  $\pm$ 0.0978 & 7.0834 $\pm$ 0.0893 & 0.1797 $\pm$ 0.0022 & 0.0012        $\pm$ 0.0002 & 0.0000 $\pm$ 0.0000
\end{tabular}
\label{tbl:logRegExperiments-baselines}
\end{table}

\begin{table}[]
\centering
\caption{Comparison of SLANG to many mean-field and full-Gaussian methods.
The performance of SLANG for different L is shown in this table, while results for mean-field and full-Gaussian methods are reported in Table \ref{tbl:logRegExperiments-baselines}.
}
\begin{tabular}{cccccc}
                                    &                         & \multicolumn{4}{c}{SLANG} \\ \cmidrule(lr){3-6}
\textbf{Datasets}                                    & \textbf{Metrics}                 & \textbf{L = 1 }& \textbf{L = 2 }& \textbf{L = 5 }& \textbf{L = 10} \\ \hline
\multirow{3}{*}{Australian}         & ELBO                    & 0.5744 $\pm$ 0.0055 & 0.5743 $\pm$ 0.0055 & 0.5690 $\pm$ 0.0056 & 0.5659 $\pm$ 0.0058\\
                                    & NLL                     & 0.3415 $\pm$ 0.0065 & 0.3416 $\pm$ 0.0065 & 0.3392 $\pm$ 0.0065 & 0.3382 $\pm$ 0.0066\\
                                    & KL {\small ($\times 10^4$)}      & 0.0332 $\pm$ 0.0068 & 0.0313 $\pm$ 0.0067 & 0.0084 $\pm$ 0.0020 & 0.0017 $\pm$ 0.0003\\ \hline
\multirow{3}{*}{
\begin{tabular}[c]{@{}l@{}}
    Breast\\ Cancer\end{tabular}}   & ELBO                    & 0.1117 $\pm$ 0.0029 & 0.1111 $\pm$ 0.0028 & 0.1114 $\pm$ 0.0028 & 0.1107 $\pm$ 0.0028\\
                                    & NLL                     & 0.0921 $\pm$ 0.0023 & 0.0918 $\pm$ 0.0023 & 0.0919 $\pm$ 0.0023 & 0.0920 $\pm$ 0.0023\\
                                    & KL                      & 0.9112 $\pm$ 0.0177 & 0.7560 $\pm$ 0.0290 & 0.8418 $\pm$ 0.0240 & 0.6376 $\pm$ 0.0222\\ \hline
\multirow{3}{*}{a1a}                & ELBO                    & 0.3766 $\pm$ 0.0000 & 0.3759 $\pm$ 0.0000 & 0.3744 $\pm$ 0.0000 & 0.3732 $\pm$ 0.0000\\
                                    & NLL                     & 0.3386 $\pm$ 0.0000 & 0.3385 $\pm$ 0.0000 & 0.3386 $\pm$ 0.0000 & 0.3386 $\pm$ 0.0000\\
                                    & KL {\small ($\times 10^2$)}      & 0.3051 $\pm$ 0.0000 & 0.2490 $\pm$ 0.0000 & 0.1731 $\pm$ 0.0000 & 0.1179 $\pm$ 0.0000\\ \hline
\multirow{3}{*}{
\begin{tabular}[c]{@{}l@{}}
    USPS\\ (3vs5)\end{tabular}}     & ELBO                    & 0.2096 $\pm$ 0.0025 & 0.2059 $\pm$ 0.0024 & 0.1979 $\pm$ 0.0024 & 0.1929 $\pm$ 0.0023\\
                                    & NLL                     & 0.1325 $\pm$ 0.0019 & 0.1325 $\pm$ 0.0019 & 0.1317 $\pm$ 0.0019 & 0.1314 $\pm$ 0.0019\\
                                    & KL ({\small $\times 10^1$)}      & 1.4924 $\pm$ 0.0199 & 1.2457 $\pm$ 0.0175 & 0.7547 $\pm$ 0.0110 & 0.4481 $\pm$ 0.0058
\end{tabular}
\label{tbl:logRegExperiments-slang}
\end{table}
\end{landscape}

\subsection{Algorithmic Details for Logistic Regression Results (Table \ref{tbl:logRegExperiments})}
\label{app:implement_logreg}

Datasets for logistic regression are available at {\footnotesize \url{https://www.csie.ntu.edu.tw/~cjlin/libsvmtools/datasets/binary.html}}.
We used the model hyper-parameters found by \cite{khan2017conjugate} for all datasets except for USPS.
All details are given in Table \ref{data_stat}, which is reproduced from their paper.
We selected a relatively strong prior for USPS to avoid overfitting, but did not search for an optimal precision.

\begin{table}[t]
\center
\caption{A list of datasets for logistic regression. $N_{\text{Train}}$ is the number of training data.
$\lambda$ is the precision of the prior distribution used in our logistic regression experiments.
}
\begin{tabular}{llllll}
\hline
Dataset & $N$ & $D$ & $N_{\text{Train}}$   & Prior Precision & $M$ \\
\hline
USPS3vs5 & 1,781 & 256 & 884 & $\lambda=25$ & 64 \\
a1a & 32,561 & 123 & 1,605  & $\lambda=2.8072$    & 128\\
Australian-scale &  690  &  14 &  345 &   $\lambda=10^{-5}$ & 32  \\
Breast-cancer-scale & 683  &  10 & 341  &   $\lambda=1.0$ & 32 \\
\hline
\end{tabular}
\label{data_stat}
\end{table}

For all datasets except a1a, we ran each method on 20 different 50\%-50\% training-test splits of the datasets.
a1a is pre-split into a training and test set and so we only report values for the provided split.
Each method was run for 10,000 epochs.
We initially set $\alpha_0 = \beta_0 = 0.05$.
We then decayed the learning rates at every iteration as follows:
\[ \alpha_t = \beta_t = \frac{\alpha_0}{(1 + t^{0.51})} \]
Using a large number of epochs and slowly decaying the learning rates ensured that the considered methods converged.
The number of MC samples used was 12.
For each dataset, we used a batch size that was roughly one-tenth of the total training set size.
These sizes are shown in Table \ref{data_stat}.
On all experiments, SLANG used momentum for the mean parameter $\vmu$, with the parameter set to $\gamma=0.9$.

Finally, we used the final covariance matrices learned on the first training-test splits of the datasets
to generate Figures \ref{fig:1} and \ref{fig:plot1-logreg}.

\begin{table}[]
    \caption{Learning rates for the logistic regression convergence experiments in Figures.}
    \centering
    \begin{tabular}{l c c c c c c c c}
                            & \multicolumn{2}{c}{Mean-Field} & \multicolumn{4}{c}{SLANG} & \multicolumn{2}{c}{Full-Gaussian}\\ \cmidrule(lr){2-3} \cmidrule(lr){4-7} \cmidrule(lr){8-9}
        \textbf{Dataset}    & \textbf{EF} & \textbf{Hess.} & \textbf{L = 1} & \textbf{L = 2} & \textbf{L = 5} & \textbf{L = 10} & \textbf{EF} & \textbf{Hess.}\\ \hline
        Australian          & 0.0215 & 0.0215 & 0.0117 & 0.0117 & 0.0117 & 0.0117 & 0.0117 & 0.0117\\
        Breast Cancer       & 0.0215 & 0.0215 & 0.0398 & 0.0398 & 0.0398 & 0.0398 & 0.0398 & 0.0398\\
        USPS 3vs5           & 0.0063 & 0.0063 & 0.0117 & 0.0117 & 0.0215 & 0.0398 & 0.0398 & 0.0398\\
    \end{tabular}
\label{table:logreg_learning_rates}
\end{table}

\subsection{Algorithmic Details for Logistic Regression Convergence Experiment (Figure \ref{fig:convergence})}

We used the same hyperparameters as in the previous logistic regression experiments on the LIBSVM datasets.
These are reported in Table \ref{data_stat}.
We used the following procedure to select the learning rates separately for each method:
\begin{enumerate}
    \item The learning rates that we considered were:
    \[ \alpha = \beta \in \{ 0.0010, 0.0018, 0.0034, 0.0063, 0.0117, 0.0215, 0.0398, 0.0736, 0.1359, 0.2512 \} \]
    \item We ran three restarts with different random seeds on the same split of the data for each potential learning rate.
    These restarts ran for 5,000 epochs with 12 MC samples and the batch sizes listed in Table \ref{data_stat}.
    We did not use a decay on the learning rate.
    \item We visually inspected the mean and variance of the training loss against epochs.
    For each method, we chose the learning rate assignment that produced the fastest convergence with tolerable
    variance. Variances were compared across methods to ensure consistency.
\end{enumerate}
The learning rates selected in this manner are reported in Table \ref{table:logreg_learning_rates}.

To obtain the final convergence results,
each method was run with ten different random seeds on the same training-test split of the datasets.
We trained for 2,000 epochs on all datasets.
The number of MC samples used was 12.
Once again, the minibatch sizes listed in Table \ref{data_stat} were used.
The learning rates were not decayed.


\section{Details for Experiments on Bayesian Neural Networks}
\label{appendix:implementation_details}

\subsection{Algorithmic Details for Regression Curves Experiment}
\label{app:toyexample}

\begin{figure*}[ttt!]
\centering
\includegraphics[width=\textwidth]{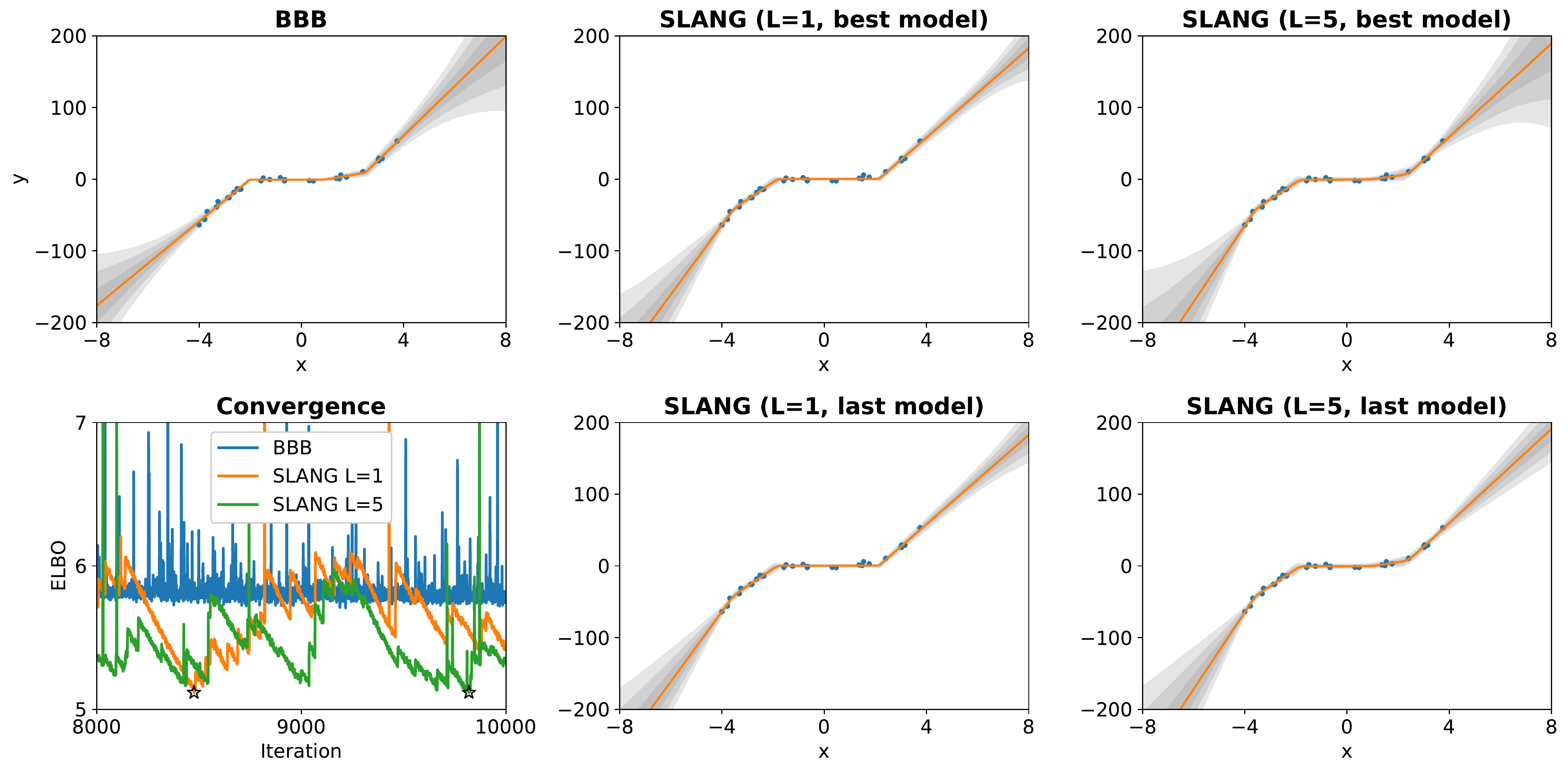}
\caption{
   Results for a synthetic toy data. Each plot shows the predictive distribution of a method along with the data examples shown in blue, except for the first plot in the bottom row which shows the value of the negative ELBO for the last 2,000 iterations.
The stars in the convergence plot indicate the selected model for SLANG-1 and SLANG-5.
}
\label{fig:covviz1d-appendix}
\end{figure*}

In Figure \ref{fig:covviz1d-appendix}, we qualitatively examine the posterior approximations computed by SLANG for neural network models using a synthetic regression data set.
The data was generated from the noisy cubic function
\[ x \sim \text{U}[-4,4] \ \text{and} \ y = x^3 + \epsilon, \ \epsilon \sim \mathcal{N}(0, 9). \]
We show the result of fitting a one-hidden-layer ReLU network with 10 units to $30$ data points generated in this manner using SLANG and BBB.
During optimization, we the used full dataset and 100 MC samples to compute stochastic gradients. We decayed the learning rates for both the mean and covariance.

All methods properly show increased uncertainty in the function when we move away from the data. In comparison to BBB, SLANG allows for smoother transitions and better representation of uncertainty at the junction of the piece-wise linear functions.

We found that the optimization procedure for SLANG did not necessarily converge on the synthetic regression dataset.
Figure \ref{fig:covviz1d-appendix} shows the value of the ELBO during the last part of the optimization procedure to illustrate the convergence issue.
This may be due to the EF matrix $\hat{\vG}(\vtheta)$ used in the VOGN update \eqref{eq:von}.
We used the ELBO to select the best model.

\subsection{Algorithmic Details for BNN Convergence Experiment}
\label{appendix:bnn_convergence}

The datasets for this experiment can be found at {\footnotesize \url{https://www.csie.ntu.edu.tw/~cjlin/libsvmtools/datasets/binary.html}}.
We used the same 50\%-50\% training-test splits of the datasets as were used in the logistic regression convergence experiment.
The models considered were feed-forward neural networks with a single hidden layer of 50 units.
The minibatch sizes were chosen to be the same those given in Table \ref{data_stat}.
We used isotropic Gaussian priors for all datasets.
On all experiments, SLANG used momentum for the mean $\vmu$ with the parameter set to $\gamma=0.9$.
Precisions for the prior distributions were chosen by grid search over the following values:
\[ \lambda \in \{ 0.001, 0.01, 0.1, 1, 8, 32, 64, 128, 512 \} \]
5-fold cross validation on the each training set was used to estimate the test log-loss;
the precisions that resulted in the smallest cross-validated test log-loss were selected.
This procedure was conducted separately for SLANG and Bayes by Backprop, but the selected values were found to agree on every dataset.
The prior precisions are listed in Table \ref{table:bnn_convergence_priors}.

\begin{table}[]
    \caption{Prior precisions for BNN convergence experiments (Figure \ref{fig:convergence}).}
    \centering
    \begin{tabular}{l c c c}
        Dataset             & \textbf{Australian} & \textbf{Breast Cancer} & \textbf{USPS 3vs5}\\ \hline
        Prior Precision     & $\lambda = 8$ & $\lambda = 8$ & $\lambda = 32$\\
    \end{tabular}
\label{table:bnn_convergence_priors}
\end{table}

We used almost the same procedure as in the logistic regression convergence experiments to select the learning rates for SLANG.
For Bayes by Backprop, we used the Adam optimizer \cite{kingma2014adam}, but carefully chose its learning rate using this procedure as well.
The procedure was as follows:
\begin{enumerate}
    \item The learning rates that we considered were:
    \begin{align*}
        \alpha = \beta \in \{ & 0.0001, 0.00021544, 0.00046416, 0.001, 0.00215443,\\
        & 0.00464159, 0.01, 0.02154435, 0.04641589, 0.1 \}
    \end{align*}
    \item We ran three restarts with different random seeds for each potential learning rate.
    These restarts ran for 1,000 epochs with 20 MC samples and the batch sizes listed in \ref{data_stat}.
    Losses were computed using 20 MC samples.
    We did not decay the learning rates.
    \item We visually inspected the mean and variance of the training loss over training epochs.
    For each method, we chose the learning rate assignment that produced the fastest convergence with tolerable
    variance. Variances were compared across methods to ensure consistency.
\end{enumerate}
The best learning rate found for SLANG was the same across each dataset and value of $L$: $\alpha = \beta = 0.02154435$.
Similarly, Bayes by Backprop performed best on each dataset with: $\alpha = 0.01$.

To obtain the final convergence results,
each method was run with ten different random seeds.
We trained for 500 epochs on all datasets.
The number of MC samples used for training and for model evaluation was 20.
Once again, the minibatch sizes listed in Table \ref{data_stat} were used.
The learning rates were not decayed.

\subsection{Algorithmic Details for UCI Experiments}
\label{appendix:uci}

Each dataset was split randomly 20 times with $90\%$ of the data in the training set and $10\%$ in the test set.
We used the same splits used in \cite{yarin16dropout}.

For both SLANG and BBB, we used an isotropic Gaussian prior
\begin{equation}
p(\vtheta|\lambda) = \mathcal{N}(\vtheta|0,\lambda^{-1}\mathrm{I})
\end{equation}
and a Gaussian likelihood
\begin{equation}
p(y|\vtheta, \vx, \tau) = \mathcal{N}(y|f(\vx, \vtheta),\tau^{-1}),
\end{equation}
where $f(\vx, \vtheta)$ is a neural network parameterized by $\vtheta$.
Following earlier work, we use 30 iterations of Bayesian optimization (BO) to tune $\lambda$ and $\tau$.
For each iteration of BO, 5-fold cross-validation was used on the current training set (for one of the 20 random splits) to evaluate the parameter setting.
For the optimal setting found by BO, one network was trained on the current training set and evaluated on the current test set.
This was repeated for each of the 20 random splits.
For each dataset, the final values reported in the table are the mean and standard error from these 20 runs.

For the 5 smallest datasets, we used a mini-batch size of 10 and 4 Monte-Carlo samples during training.
For the 3 larger datasets, we used a mini-batch size of 100 and 2 Monte-Carlo samples during training.
For all runs we used 120 epochs for the methods to converge.

For BBB, we used the Adam \cite{kingma2014adam} optimizer.
The learning rates were set individually for each method and dataset based on the cross-validation performances from an initial random search over learning rates, prior precision and noise precisions. The random search was also used to determine the search spaces for the Bayesian optimization used to tune the prior precision and the noise precision.

\subsection{Algorithmic Details for MNIST Experiments}
\label{appendix:mnist}

We fit a Bayesian neural network with two hidden layers, each with 400 hidden units and ReLU activations, to the MNIST dataset. This was done using SLANG with $L \in \{1,2,4,8,16,32\}$.
During training, a batch size of 200 was used along with 4 MC samples.
The momentum was set to $\gamma=0.9$.
The learning rates $\alpha$ and $\beta$ were initialized to $\alpha_0 = 0.1$ and decayed according to
\[ \alpha_t = \beta_t = \frac{\alpha_0}{(1 + t^{\omega})}, \]
where $\omega$ is the decay rate.

The prior precision $\lambda$, along with the learning rate decay rate $\omega$ was tuned. Denote $\sigma := \sqrt{1/\lambda}$.
For each value of $L$, we considered each combination of
\[ (- \log \sigma, \omega) \in \{0,1,2\} \times \{0.52, 0.54, 0.56, 0.58, 0.60\}. \]
The 60,000 training points for MNIST were split into a training set of 50,000 and a validation set of 10,000.
After training for 100 epochs, the best performing configuration, according to validation error, was selected for each value of $L$.
The models selected by the tuning procedure were trained further on all 60,000 training points for 300 more epochs.
Finally, each model made predictions on the test set. Both during computation of the validation and the test loss, 1,000 MC samples were used.

\end{document}